\documentclass[runningheads]{llncs}
\usepackage{graphicx}
\usepackage{comment}
\usepackage{amsmath,amssymb} 
\usepackage{color}
\usepackage{tabu}
\usepackage{float}
\usepackage{subfigure}
\usepackage{wrapfig}
\usepackage{multirow}
\usepackage{array}
\usepackage{enumerate}
\usepackage{booktabs}
\usepackage{epstopdf}
\usepackage{algorithm}
\usepackage{algorithmic}

\newcommand{\tabincell}[2]{\begin{tabular}{@{}#1@{}}#2\end{tabular}}  

\begin{document}
\pagestyle{headings}
\mainmatter
\def\ECCVSubNumber{2426}  

\title{Interpretable Neural Network Decoupling} 


\titlerunning{Interpretable Neural Network Decoupling}
%
\author{
	Yuchao Li$^{1}$,
	Rongrong Ji$^{1,2}$\thanks{Corresponding author.}
,
    Shaohui Lin$^{3}$,
    Baochang Zhang$^{4}$, \\
    Chenqian Yan$^{1}$,
    Yongjian Wu$^{5}$,
    Feiyue Huang$^{5}$,
    Ling Shao$^{6,7}$
    }
\authorrunning{Li et al.}
%
\institute{
$^{1}$Department of Artificial Intelligence, School of Informatics, Xiamen University, China,
$^{2}$Peng Cheng Laboratory, Shenzhen, China,
$^{3}$National University of Singapore, Singapore,
$^{4}$Beihang University, China,
$^{5}$BestImage, Tencent Technology (Shanghai) Co.,Ltd, China,
$^{6}$Mohamed bin Zayed University of Artificial Intelligence, Abu Dhabi, UAE,
$^{7}$Inception Institute of Artificial Intelligence, Abu Dhabi, UAE
}
\maketitle

\begin{abstract}
The remarkable performance of convolutional neural networks (CNNs) is entangled with their huge number of uninterpretable parameters, which has become the bottleneck limiting the exploitation of their full potential.
Towards network interpretation, previous endeavors mainly resort to the single filter analysis, which however ignores the relationship between filters.
In this paper, we propose a novel architecture decoupling method to interpret the network from a perspective of investigating its calculation paths.
More specifically, we introduce a novel architecture controlling module in each layer to encode the network architecture by a vector.
By maximizing the mutual information between the vectors and input images, the module is trained to select specific filters to distill a unique calculation path for each input.
Furthermore, to improve the interpretability and compactness of the decoupled network, the output of each layer is encoded to align the architecture encoding vector with the constraint of sparsity regularization.
Unlike conventional pixel-level or filter-level network interpretation methods, we propose a path-level analysis to explore the relationship between the combination of filter and semantic concepts, which is more suitable to interpret the working rationale of the decoupled network.
Extensive experiments show that the decoupled network achieves several applications, i.e., network interpretation, network acceleration, and adversarial samples detection.
%
%
\end{abstract}

\section{Introduction}

Deep convolutional neural networks (CNNs) have dominated various computer vision tasks, such as object classification, detection and semantic segmentation.
However, the superior performance of CNNs is rooted in their complex architectures and huge amounts of parameter, which thereby restrict the interpretation of their internal working mechanisms.
Such a contradiction has become a key drawback when the network is used in task-critical applications such as medical diagnosis, automatic robots, and self-driving cars.


%
To this end, network interpretation have been explored to improve the understanding of the intrinsic structures and working mechanisms of neural networks \cite{zeiler2014visualizing,bau2017network,morcos2018importance,zhang2018interpretable,koh2017understanding,lundberg2017unified,chen2019explaining}.
Interpreting a neural network involves investigating the rationale behind the decision-making process and the roles of its parameters.
For instance, some methods \cite{lakkaraju2017identifying,chen2019explaining} view networks as a whole when explaining their working process.
However, these approaches are too coarse-grained for exploring the intrinsic properties in the networks.
In contrast, network visualization approaches \cite{zeiler2014visualizing,yosinski2015understanding} interpret the role of each parameter by analyzing the pixel-level feature representation, which always require complex trial-and-error experiments.
Beyonds, Bau \emph{et al.} \cite{bau2017network} and Zhang \emph{et al.} \cite{zhang2018interpretable} explored the different roles of filters in the decision-making process of a network.
Although these methods are more suitable for explaining the network, they characterize semantic concepts using only a single filter, which has been proven to be less effective than using a combination of multiple filters \cite{wang2015unsupervised,fong2018net2vec}.
Under this situation, different combination of filters can be viewed as different calculation paths in the network, which inspires us to investigate the working process of networks based on a path-level analysis.
The challenge, however, comes from the fact that each inference involves all filters in the network and has the same calculation process, making it difficult to interpret how each calculation path affects the final result.
To overcome this problem, previous methods \cite{wang2018interpret,sun2019adaptive} explore the difference between the calculation paths of different inputs by reducing the number of parameters involved in the calculation process.
For instance, Wang \emph{et al.} \cite{wang2018interpret} proposed a post-hoc analysis to obtain a unique calculation path of a specific input based on a pre-trained model, which however involves a huge number of complicated experiments.
Moreover, Sun \emph{et al.} \cite{sun2019adaptive} learned a network that generates a dynamic calculation path in the last layer by modifying the SGD algorithm.
However, it ignores the fact that the responses of filters are also dynamic in the intermediate layers, and thus cannot interpret how the entire network works.

\begin{wrapfigure}{r}{4cm}
  \vspace{-2em}
  \hspace{0.2em}
\includegraphics[width=0.3\textwidth]{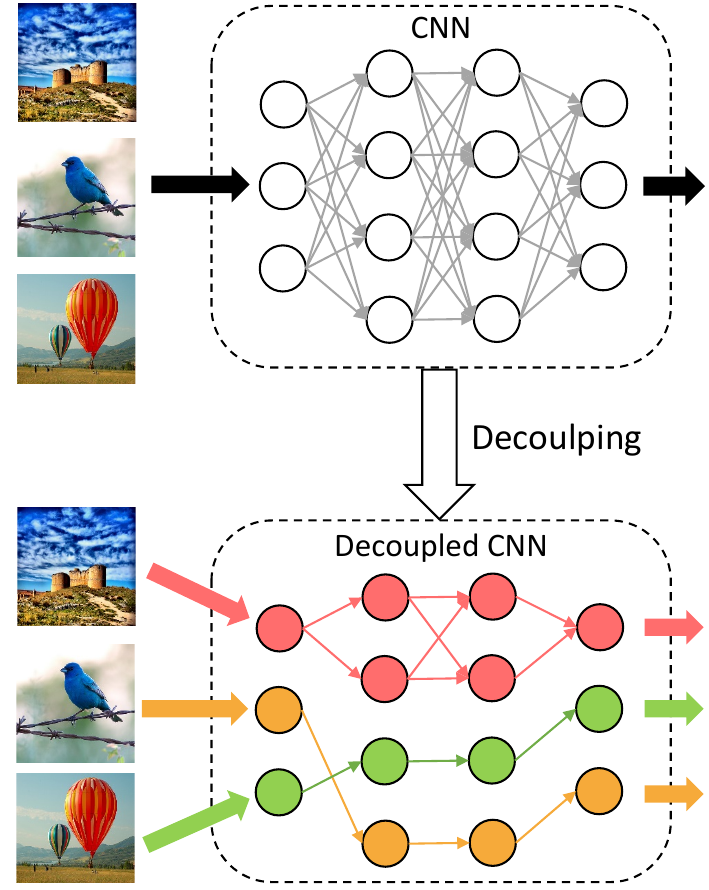}
\caption{An example of the neural network architecture decoupling. Each color represents a calculation path of specific input.}
\label{fig:ad}
\vspace{-1em}
\end{wrapfigure}
\vspace{-0.1em}
In this paper, we propose an interpretable network decoupling approach, which enables a network to adaptively select a suitable subset of filters to form a calculation path for each input, as shown in Fig.~\ref{fig:ad}. 
In particular, Our design principle lies in a novel light-weight \emph{architecture controlling module} as well as a novel learning process for network decoupling. 
Fig.~\ref{fig:framework} depicts the framework of the proposed method.
The architecture controlling module is first incorporated into each layer to dynamically select filters during network inference with a negligible computational burden.
Then, we maximize the mutual information between the architecture encoding vector (\emph{i.e.,} the output of the architecture controlling module) and the inherent attributes of the input images during training, which allows the network to dynamically generate the calculation path related to the input.
In addition, to further improve the interpretability of decoupled networks, we increase the similarity between the architecture encoding vector of each convolutional layer and its output by minimizing the KL-divergence between them, making filter only respond to a specific object.
Finally, we sparsify the architecture encoding vector to attenuate the calculation path and eliminate the effects of redundant filters for each input.
We also introduce an improved semantic hashing scheme to make the discrete architecture encoding vector differentiable, which is therefore capable to be trained directly by stochastic gradient descent (SGD).
\begin{figure}[t]
    \label{fig:framework}
  \centering
    \includegraphics[width=1\columnwidth]{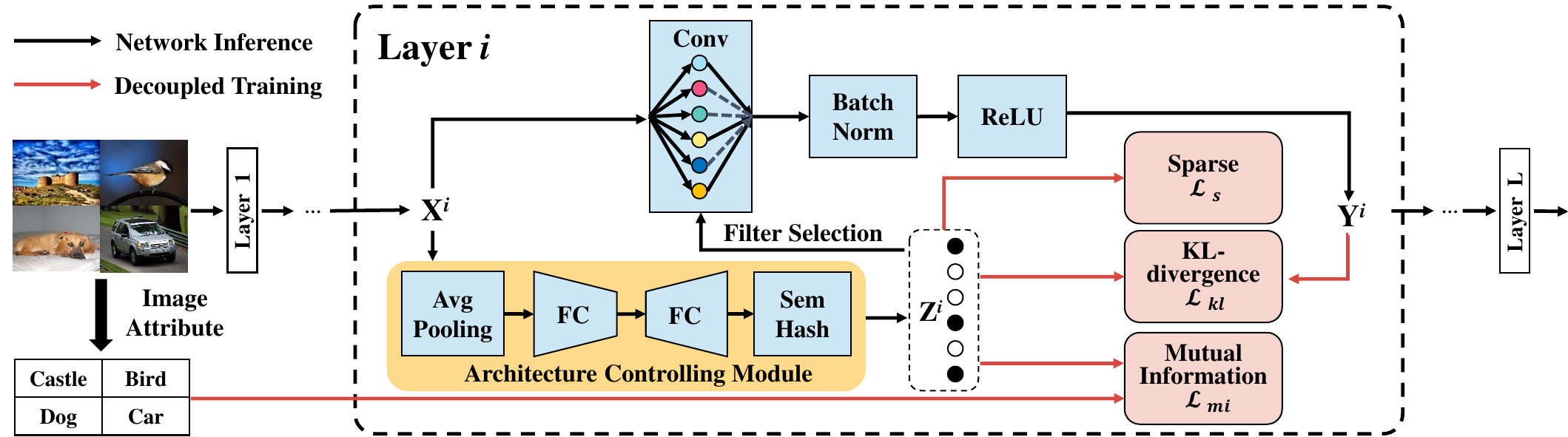}
  \vspace{-2em}
     \caption{The framework of the proposed interpretable neural network decoupling. The architecture encoding vector $\mathbf{z}^i$ is first constructed by the architecture controlling module, and then learned to determine the filter selection by Eq.~\ref{eq:all}. For network inference, we only use the selected filters based on each input. The mutual information loss $\mathcal{L}_{mi}$ is computed between the output of the architecture controlling module $\mathbf{z}^i$ and the attribute of the inputs to decouple the network architecture. The KL-divergence loss $\mathcal{L}_{kl}$ is computed by the output of convolutional layer $\mathcal{Y}^i$ and $\mathbf{z}^i$ to disentangle the filters. The sparse loss $\mathcal{L}_{s}$ is used to sparsify the result of filter selection.}
     \vspace{-1em}
  \label{fig:framework}
\end{figure}

Correspondingly, the decoupled network becomes more interpretable, and one can trace the functional processing behavior layer-by-layer to form a hierarchical path towards understanding the working principle of the decoupled network.
Meanwhile, each filter is only related to a set of similar input images after the decoupling, thus they also become more interpretable, and the combination of them forms a decoupled sub-architecture, which better characterizes the specific semantic concepts.
Such a decoupled architecture further benefits from a low computational cost for network acceleration, as well as good hints for adversarial samples detection, which are subsequently validated in our experiments.

We summarize our three main contributions as follows:
\vspace{-1em}
\begin{itemize}
\item
To interpret neural networks by dynamically selecting the filters for different inputs, we propose a lightweight architecture controlling module, which is differentiable and can be optimized by SGD based on the losses we propose.

\item
The decoupled network reserves similar performance of the original network and has better interpretable. 
Thus it enables the functional processing of each calculation path to be well interpreted, which helps better understand the rationale behind the network inference, as well as explore the relationship between filters and semantic concepts in the decoupled network.

\item
Our method is generic and flexible, which can be easily employed on the existing network architectures, such as VGGNets \cite{simonyan2014very}, ResNets \cite{he2016deep}, and Inceptions \cite{szegedy2015going}.
The decoupled architecture further benefits extensive applications, including network acceleration and adversarial samples detection.
\end{itemize}

\vspace{-1em}
\section{Related Work}

\textbf{Network Interpretation.}
One way to interpret a network is to analyze how it responds to a specific input image for output prediction \cite{koh2017understanding,lundberg2017unified,lakkaraju2017identifying,zhang2019interpreting,chen2019explaining}.
This strategy views the network as a whole to interpret the network prediction results by exploring the knowledge blind spots of neural networks \cite{lakkaraju2017identifying}, or by assigning each output feature an importance value for a particular prediction \cite{lundberg2017unified}. 
Moreover, a decision tree \cite{zhang2019interpreting} or an explainer network \cite{chen2019explaining} has been used to better understand the classification process.
However, these methods only pay attention to the reason behind the network prediction result, and the roles of each parameter are ignored, making it difficult to understand their effects on the network.

To open the black-box of neural network and interpret the role of parameters, several methods \cite{zeiler2014visualizing,yosinski2015understanding,dosovitskiy2016inverting} have been proposed to visualize the feature representations inside the network.
For instance, Zeiler \emph{et al.} \cite{zeiler2014visualizing} visualized the feature maps in the intermediate layers by establishing a deconvolutional network corresponding to the original one.
Yoshinski \emph{et al.} \cite{yosinski2015understanding} proposed two visualization methods to explore the information contained in features: a respective post-hoc analysis on a pre-trained model and learning a network by regularized optimization.
Visualizing feature representations is a very direct method to explain the role of parameters in a network, which however requires extensive experiments due to the enormous number of parameters.

In addition to the above methods, the functions of filters are also explored for interpreting networks \cite{yosinski2014transferable,bau2017network,morcos2018importance,zhang2018interpretable,sun2019adaptive}.
They have evaluated the transferability of filters \cite{yosinski2014transferable} or quantified the relationship between filters and categories \cite{morcos2018importance} to explain their different roles.
Compared with using a single filter to represent semantic concepts, methods in \cite{wang2015unsupervised,fong2018net2vec} have found that the semantic concepts can be better characterized by combining multiple filters.
Wang \emph{et al.} \cite{wang2015unsupervised} further validated that clustering the activations of multiple filters can better represent semantic concepts than using a single filter.
Fong \emph{et al.} \cite{fong2018net2vec} mapped the semantic concepts into vectorial embeddings based on the responses of multiple filters and found that these embeddings can better characterize the features. 
Different from these methods, we interpret the working principle of a network based on a path-level analysis by decoupling the network, upon which we further disentangle each intra-layer filter to explore the interpretable semantic concepts across filters on the calculation path.
Our method is more in line with the internal working mechanism of the network than these works, and has a better extension to other applications, such as network acceleration and adversarial samples detection.

\textbf{Conditional Computation.}
Works on conditional computation tend to concentrate on the selection of model components when generating the calculation path. 
For instance, the work in \cite{bengio2013estimating} explored the influence of stochastic or non-smooth neurons when estimating the gradient of the loss function.
Later, an expert network was learned to find a suitable calculation path for each input by reinforcement learning \cite{bolukbasi2017adaptive} or SGD \cite{chen2018gaternet}.
However, the requirement of a specific expert network makes these approaches cumbersome.
Along another line, a halting score \cite{figurnov2017spatially} or a differentiable directed acyclic graph \cite{liu2018dynamic} has been used to dynamically adjust the model components involved in the calculation process.
Recently, a feature boosting and suppression method \cite{gao2018dynamic} was introduced to skip unimportant output channels of the convolutional layer at runtime.
However, it selects the same number of filters for each layer, without considering inter-layer differences.
Different from the above works, we employ a novel architecture controlling module to decouple the network by fitting it to the data distribution.
After decoupling, the network becomes interpretable, enabling us to visualize its intrinsic structure, accelerate the inference, and detect adversarial samples.

\vspace{-0.5em}
\section{Architecture Decoupling}
Formally speaking, the $l$-th convolutional layer in a network with a batch normalization (BN) \cite{ioffe2015batch} and a ReLU layer \cite{nair2010rectified} transforms $\mathcal{X}^l \in \mathbb{R}^{C^{l}\times H_{in}^{l}\times W_{in}^{l}}$ to $\mathcal{Y}^l \in \mathbb{R}^{N^l\times H_{out}^l\times W_{out}^l}$ using the weight $\mathcal{W}^l \in \mathbb{R}^{N^l \times C^l \times D^l \times D^l}$, which is defined as:
\begin{equation}
  \small
  \label{eq:Y}
\mathcal{Y}^l = \Big(BN \big (Conv(\mathcal{X}^l, \mathcal{W}^l) \big) \Big)_{+},
\end{equation}
where $(\cdot)_{+}$ represents the ReLU layer, and Conv($\cdot, \cdot$) denotes the standard convolution operator.
$(H_{in}^{l}, W_{in}^{l})$ and $(H_{out}^{l}, W_{out}^{l})$ are the spatial size of the input and output in the $l$-th layer, respectively.
$D^l$ is the kernel size.

\subsection{Architecture Controlling Module}
For an input image, the proposed architecture controlling module selects the filters and generates the calculation path during network inference.
In particular, we aim to predict which filters need to participate in the convolutional computation \textit{before} the convolutional operation to accelerate network inference.
Therefore, for the $l$-th convolutional layer, the architecture encoding vector $\mathbf{z}^l$ (\emph{i.e.,} the output of the architecture controlling module) only relies on the input $\mathcal{X}^l$ instead of the output $\mathcal{Y}^l$, which is defined as $\mathbf{z}^l = G^l(\mathcal{X}^l)$.
Inspired by the effectiveness of the squeeze-and-excitation (SE) block \cite{hu2018squeeze}, we select a similar SE-block to predict the importance of each filter.
Thus, we first squeeze the global spatial information via global average pooling, which transforms each input channel $X_i^l\in \mathbb{R}^{H_{in}^{l}\times W_{in}^{l}}$ to a scalar $s_i^l$. 
We then design a sub-network structure $\bar{G}^l(\mathbf{s}^l)$ to determine the filter selection based on $\mathbf{s}^l \in \mathbb{R}^{C^l}$, which is formed by two fully connected layers, \emph{i.e.,} a dimensionality-reduction layer with weights $\mathbf{W}^l_1$ and a dimensionality-increasing layer with weights $\mathbf{W}^l_2$:
\begin{equation}
  \small
  \label{eq:G}
\bar{G}^l(\mathbf{s}^l) = \mathbf{W}^l_2 \cdot (\mathbf{W}^l_1 \cdot \mathbf{s}^l)_{+},
\end{equation}
where $\mathbf{W}^l_1 \in \mathbb{R}^{\frac{C^l}{\gamma}\times C^{l}}$, $\mathbf{W}^l_2 \in \mathbb{R}^{N^l\times \frac{C^l}{\gamma}}$ and $\cdot$ represents the matrix multiplication.
We ignore the bias for simplicity.
To reduce the module complexity, we empirically set the reduction ratio $\gamma$ to $4$ in our experiments.
The output of $\bar{G}^l(\mathbf{s}^l)$ is a real vector, while we need to binarize $\bar{G}^l(\mathbf{s}^l)$ to construct a binary vector $\mathbf{z}^l$, which represents the result of filter selection.
However, a simple discretization using the sign function is not differentiable, which prevents the corresponding gradients from being directly obtained by back-propagation.
Thus, we further employ an \textit{Improved SemHash} method \cite{kaiser2018fast} to transform the real vector in $\bar{G}^l(\mathbf{s}^l)$ to a binary vector by a simple rounding bottleneck, which also makes the discretization become differentiable.

\textbf{Improved SemHash.}
The proposed scheme is based on the different operations for training and testing.
During training, we first sample a noise $\alpha \sim\mathcal{N}(0,1)^{N^l}$, which is added to $\bar{G}^l(\mathbf{s}^l)$, and then obtain $\widetilde{\mathbf{s}}^l = \bar{G}^l(\mathbf{s}^l) + \alpha$.
After that, we compute a real vector and a binary vector by:
\begin{equation}
  \small
\mathbf{v}^l_1 = \sigma'(\widetilde{\mathbf{s}}^l), \mathbf{v}^l_2 = \textbf{1}(\widetilde{\mathbf{s}}^l>0),
\end{equation}
where $\sigma'$ is a saturating Sigmoid function \cite{kaiser2016can} denoted as:
\begin{equation}
  \small
\sigma'(x) = \max \Big(0, \min \big(1, 1.2\sigma(x) - 0.1 \big) \Big).
\end{equation}
Here, $\sigma$ is the Sigmoid function.
$\mathbf{v}^l_1 \in \mathbb{R}^{C^l}$ is a real vector with all elements falling in the interval $[0, 1]$, and we calculate its gradient during back-propagation.
$\mathbf{v}^l_2 \in \mathbb{R}^{C^l}$ represents the discretized vector, which cannot be involved in the gradient calculation.
Thus, we randomly use $\mathbf{z}^l = \mathbf{v}^l_1$ for half of the training samples and $\mathbf{z}^l = \mathbf{v}^l_2$ for the rest in the forward-propagation.
We then mask the output channels using the architecture encoding vector (\emph{i.e.,} $\mathcal{Y}^l * \mathbf{z}^l$) as the final output of this layer.
In the backward-propagation, the gradient of $\mathbf{z}^l$ is the same as the gradient of $\mathbf{v}^l_1$.
During evaluation/testing, we directly use the sign function in the forward-propagation as:
\begin{equation}
  \small
  \label{eq:5}
\mathbf{z}^l = \textbf{1} \big(\bar{G}^l(\mathbf{s}^l)>0 \big).
\end{equation}

After that, we select suitable filters involved in the convolutional computation based on $\mathbf{z}^l$ to achieve fast inference.

\vspace{-1em}
\subsection{Network Training}
\label{sec:3.3}
We expect the network architecture to be gradually decoupled during training, where the essential problem is how to learn an architecture encoding vector that fits the data distribution.
To this end, we propose three loss functions for network decoupling.

\textbf{Mutual Information Loss.}
When the network architecture is decoupled, different inputs should select their related sets of filters.
We adopt mutual information $I(a;\mathbf{z}^l)$ between the result of filter selection $\mathbf{z}^l$ and the attribute of an input image $a$ (\emph{i.e.,} the unique information contained in the input image) to measure the correlation between the architecture encoding vector and its input image.
$I(a;\mathbf{z}^l)=0$ means that the result of filter selection is independent to the input image, \emph{i.e.,} all the inputs share the same filter selection.
In contrast, when $I(a;\mathbf{z}^l) \neq 0$, filter selection depends on the input image.
Thus, we maximize the mutual information between $a$ and $\mathbf{z}^l$ to achieve architecture decoupling.
Formally speaking, we have:
\vspace{-0.5em}
\begin{equation}
  \small
    \begin{aligned}
I(a;\mathbf{z}^l) &= H(a) - H(a|\mathbf{z}^l) \\
    &= \sum_{a} \sum_{\mathbf{z}^l} P(a, \mathbf{z}^l) logP(a|\mathbf{z}^l) + H(a) \\
    &= \sum_{a} \sum_{\mathbf{z}^l} P(\mathbf{z}^l) P(a|\mathbf{z}^l) log P(a|\mathbf{z}^l) + H(a).
    \end{aligned}
    \vspace{-0.5em}
\end{equation}

The mutual information $I(a;\mathbf{z}^l)$ is difficult to directly maximize, as it is hard to obtain $P(a|\mathbf{z}^l)$.
Thus, we use $Q(a|\mathbf{z}^l)$ as a variational approximation to $P(a|\mathbf{z}^l)$ \cite{agakov2004algorithm}.
In fact, the KL-divergence is positive, so we have:
\begin{equation}
  \small
    \begin{aligned}
KL \big( P(a|\mathbf{z}^l), Q(a|\mathbf{z}^l) \big) \ge 0 &\Rightarrow \sum_{a} P(a|\mathbf{z}^l) log P(a|\mathbf{z}^l) \\
  &\ge \sum_{a} P(a|\mathbf{z}^l) log Q(a|\mathbf{z}^l).
    \end{aligned}
\end{equation}
We then obtain the following equation:
\begin{equation}
  \label{eq:8}
  \small
    \begin{aligned}
I(a;\mathbf{z}^l) &\ge \sum_{a} \sum_{\mathbf{z}^l} P(\mathbf{z}^l) P(a|\mathbf{z}^l) log Q(a|\mathbf{z}^l) + H(a) \\
         &\ge \sum_{a} \sum_{\mathbf{z}^l} P(\mathbf{z}^l) P(a|\mathbf{z}^l) log Q(a|\mathbf{z}^l) \\
         &=\mathbb{E}_{\mathbf{z}^l \sim G^l(\mathcal{X}^l)}[\mathbb{E}_{a \sim P(a|\mathbf{z}^l)}[logQ(a|\mathbf{z}^l)]].
    \end{aligned}
\end{equation}
Eq.~\ref{eq:8} provides a lower bound for the mutual information $I(a;|\mathbf{z}^l)$.
By maximizing this bound, the mutual information $I(a;\mathbf{z}^l)$ will also be maximized accordingly.
In our paper, we use the class label as the attribute of the input image $c$ in the classification task.
Moreover, we reparametrize $Q(a|\mathbf{z}^l)$ as a neural network $\tilde{Q}(\mathbf{z}^l)$ that contains a fully connected layer and a softmax layer.
Thus, maximizing the mutual information in Eq.~\ref{eq:8} is achieved by minimizing the following loss:
\vspace{-0.5em}
\begin{equation}
  \small
  \label{eq:mi}
\mathcal{L}_{mi} = -\sum_{l=1}^{L} A_X * log \tilde{Q}(\mathbf{z}^l),
\vspace{-0.5em}
\end{equation}
where $A_X$ represents the label of the input image $X$.
$\tilde{Q}(\mathbf{z}^l)$ is defined as $\mathbf{W}^l_{cla} \cdot \mathbf{z}^l$ with a fully connected weight $\mathbf{W}^l_{cla} \in \mathbb{R}^{K \times N^l}$, where $K$ represents the number of categories  in image classification.

\textbf{KL-divergence Loss.}
After decoupling the network architecture, we guarantee that the filter selection depends on the input image.
However, it is uncertain whether the filters become different (\emph{i.e.,} detect different objects), which obstructs us from further interpreting the network.
If a filter only responds to a specific semantic concept, it will not be activated when the input does not contain this feature.
Thus, by limiting filters to only respond to specific category, they can be disentangled to detect different categories.
To achieve this goal, we minimize the KL-divergence between the output of the current layer and its corresponding architecture encoding vector, which ensures that the overall responses of filters have a similar distribution to the responses of the selected subset.
To align the dimension of the convolution output and architecture encoding vector, we further downsample $\mathcal{Y}^l$ to $\mathbf{y}^l \in \mathbb{R}^{N}$ using global average pooling.
Then, the KL-divergence loss is defined as:

  \vspace{-1em}
\begin{equation}
  \small
  \label{eq:kl}
\mathcal{L}_{kl} = \sum_{l=1}^{L} KL(\mathbf{z}^l || \mathbf{y}^l).
\end{equation}

As the output of filter is limited by the result of filter selection, it will be unique and only detects the specific object.
Thus, all filters are different from each other, \emph{i.e.,} each one performs its function.

\textbf{Sparse Loss.}
An $\ell_1$-regularization on $\mathbf{z}^l$ is further introduced to encourage the architecture encoding vector to be sparse, which makes the calculation path of each input becomes thinner.
Thus, the sparse loss is defined as:

\begin{equation}
  \small
  \label{eq:s}
\mathcal{L}_{s} = \sum_{l=1}^{L} | \| \mathbf{z}^l \|_1 - R*N^l|,
\end{equation}
where $R$ represents the target compression ratio.
Since $z^l$ falls in the interval $[0, 1]$, the maximum value of $\| \mathbf{z}^l \|_1$ is $N^l$, and the minimum value is $0$, where $N^l$ is the number of filters.
For example, we set $R$ to 0.5 if activating only half of the filters.

Therefore, we obtain the overall loss function as follows:
\begin{equation}
  \small
  \label{eq:all}
\mathcal{L} = \mathcal{L}_{ce} + \lambda_m * \mathcal{L}_{mi} + \lambda_k * \mathcal{L}_{kl} + \lambda_s * \mathcal{L}_{s},
\end{equation}
where $\mathcal{L}_{ce}$ is the network classification loss.
$\lambda_m$, $\lambda_k$ and $\lambda_s$ are the hyper-parameters.
Eq.~\ref{eq:all} can be effectively solved via SGD.

%
\vspace{-0.5em}
\section{Experiments}
\vspace{-0.5em}
We evaluate the effectiveness of the proposed neural network architecture decoupling scheme on three kinds of networks, \emph{i.e.}, VGGNets \cite{simonyan2014very}, ResNets \cite{he2016deep}, and Inceptions \cite{szegedy2015going}.
For network acceleration, we conduct comprehensive experiments on three datasets, \emph{i.e.}, CIFAR-10, CIFAR-100 \cite{krizhevsky2009learning} and ImageNet 2012 \cite{russakovsky2015imagenet}.
CIFAR-10 and CIFAR-100 contain 50,000 training images and 10,000 testing images from 10 and 100 classes, respectively.
%
ImageNet 2012 consists of 1.28 million training images and 50,000 validation images from 1,000 classe.
For quantifying the network interpretability, we use the interpretability of filters \cite{zhang2018interpretable} and the representation ability of semantic features \cite{fong2018net2vec} on BRODEN dataset \cite{bau2017network} to evaluate the original and our decoupled models.
BRODEN contains over 60,000 images with pixel-level and image-level annotations for 1,197 concepts across 6 categories: scenes, objects, parts, materials, textures, and colors. 
The indicator of interpretability is detailed discussed in Section 4.2.

\subsection{Implementation Details}
We implement our method using PyTorch \cite{paszke2017automatic}.
The weights of decoupled networks are initialized using the weights from their corresponding pre-trained models.
We add the architecture controlling module to all convolutional layers except the first and last ones.
All networks are trained using stochastic gradient descent with a momentum of 0.9.
For CIFAR-10 and CIFAR-100, we train all the networks over 200 epochs using a mini-batch size of 128.
The learning rate is initialized by 0.1, which is divided by 10 at 50\% and 75\% of the total number of epochs.
For ImageNet 2012, we train the networks over 120 epochs with a mini-batch size of 64 and 256 for VGG-16 and ResNet-18, respectively.
The learning rate is initialized as 0.01 and is multiplied by 0.1 after the $30$-th, $60$-th and $90$-th epoch.
The real speed on the CPU is measured by a single-thread AMD Ryzen Threadripper 1900X.
%
Except for the experiments on network acceleration, we automatically learn sparse filters by setting $R$ to 0 in Eq.~\ref{eq:s}.

\begin{figure}[t]
  \centering
    \subfigure[VGGNet] {
    \label{fig:arch_vggnet}
    \includegraphics[width=0.3\columnwidth]{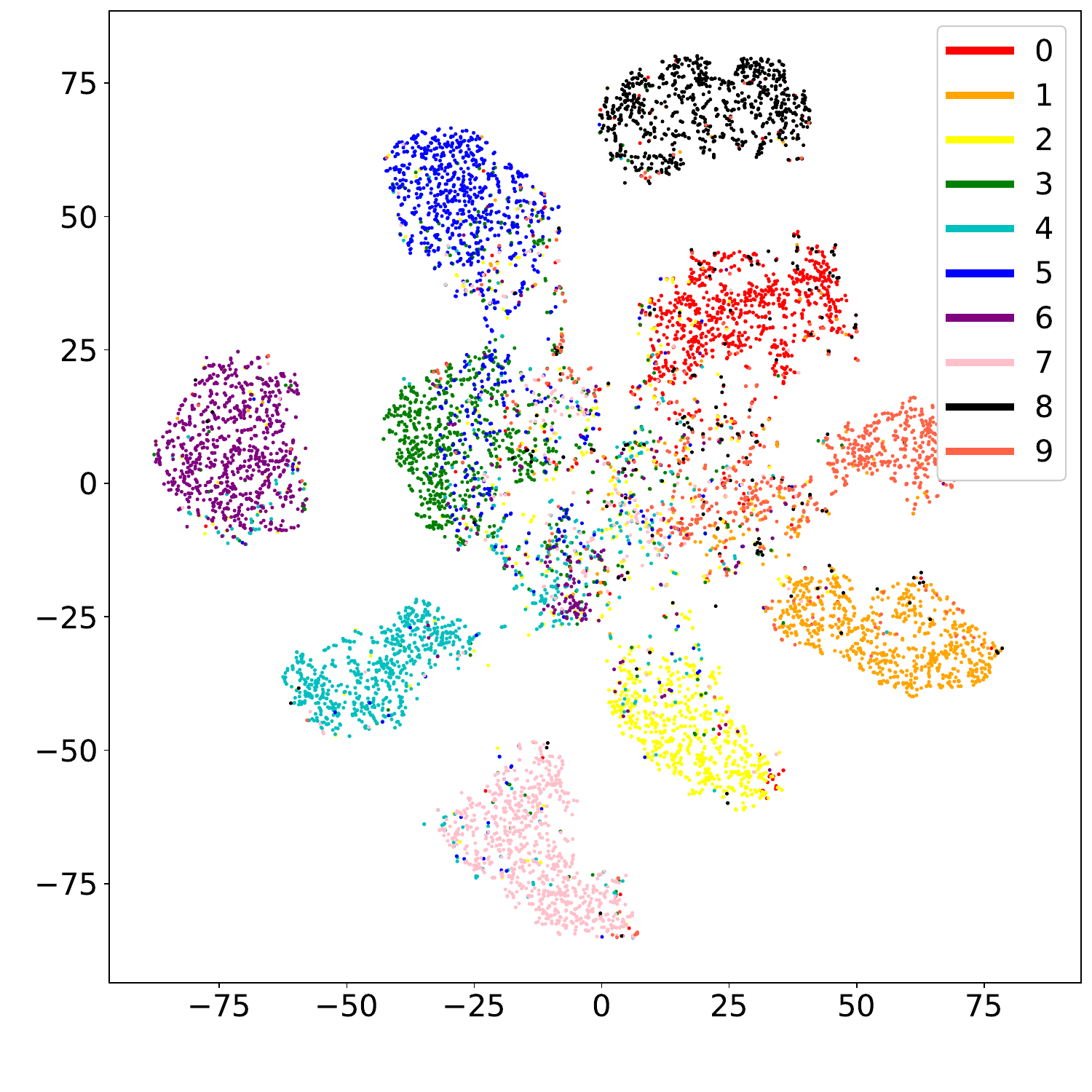}
  }
  \hspace{-1em}
  \subfigure[ResNet-56] {
  \label{fig:arch_resnet56}
  \includegraphics[width=0.3\columnwidth]{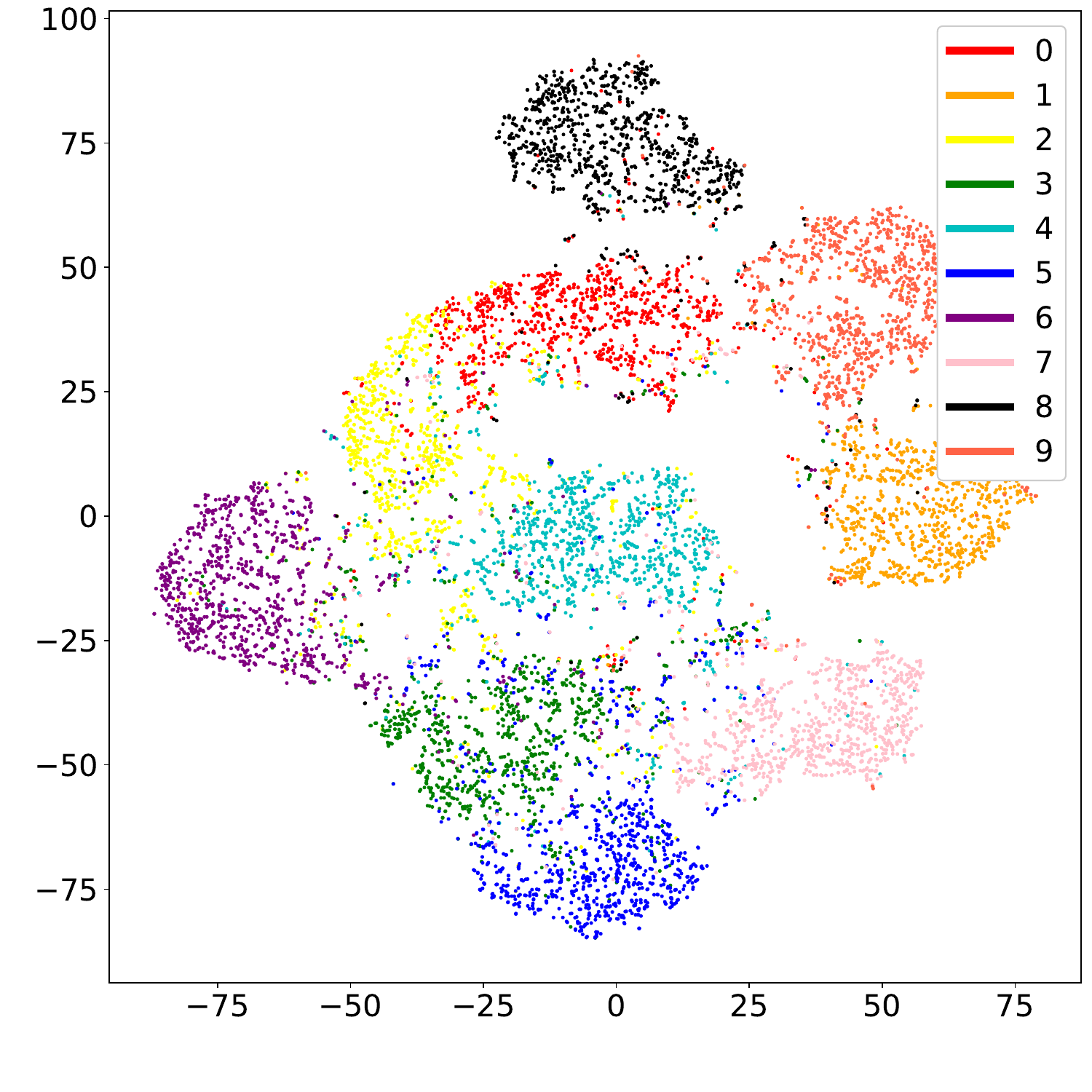}
  }
  \hspace{-1em}
  \subfigure[GoogleNet] {
  \label{fig:arch_googlenet}
  \includegraphics[width=0.3\columnwidth]{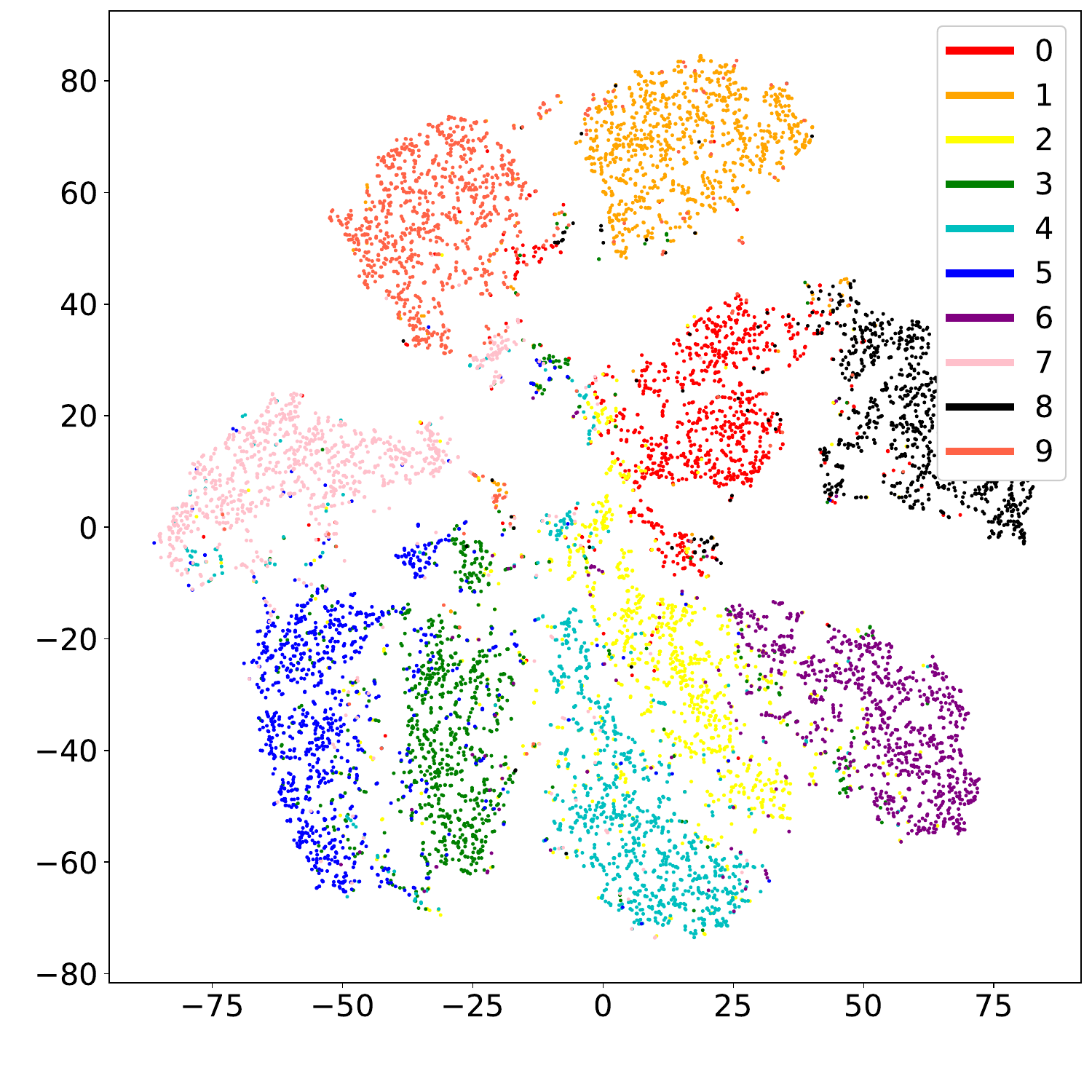}
  }
  \vspace{-1em}
     \caption{Visualization of the distribution of the integral calculation path in different networks on CIFAR-10.}
     \vspace{-1em}
  \label{fig:arch_draw}
  \end{figure} 

\vspace{-1em}
\subsection{Network Interpretability}
\label{sec:net_inter}

\textbf{Architecture Encoding.}
We collect the calculation paths from three different networks (\emph{i.e.,} VGGNet, ResNet-56 and GoogleNet) to verify that the proposed network decoupling method can successfully decouple the network and ensure that it generates different calculation paths for different images.
We first reduce the dimension of the calculation path (\emph{i.e.,} the concatenation of architecture encoding vectors $\mathbf{z}^l$ across all layers) to 300 using Principal Component Analysis (PCA), and then visualize the calculation path by t-SNE \cite{maaten2008visualizing}.
As shown in Fig.~\ref{fig:arch_draw}, each color represents one category and each dot is a calculation path corresponding to an input.
We can see that the network architecture is successfully decoupled after training by our method, where different categories of images have different calculation paths.

\begin{wrapfigure}{r}{5.5cm}
  \vspace{-1.5em}
  \hspace{0.5em}
\includegraphics[width=0.45\textwidth]{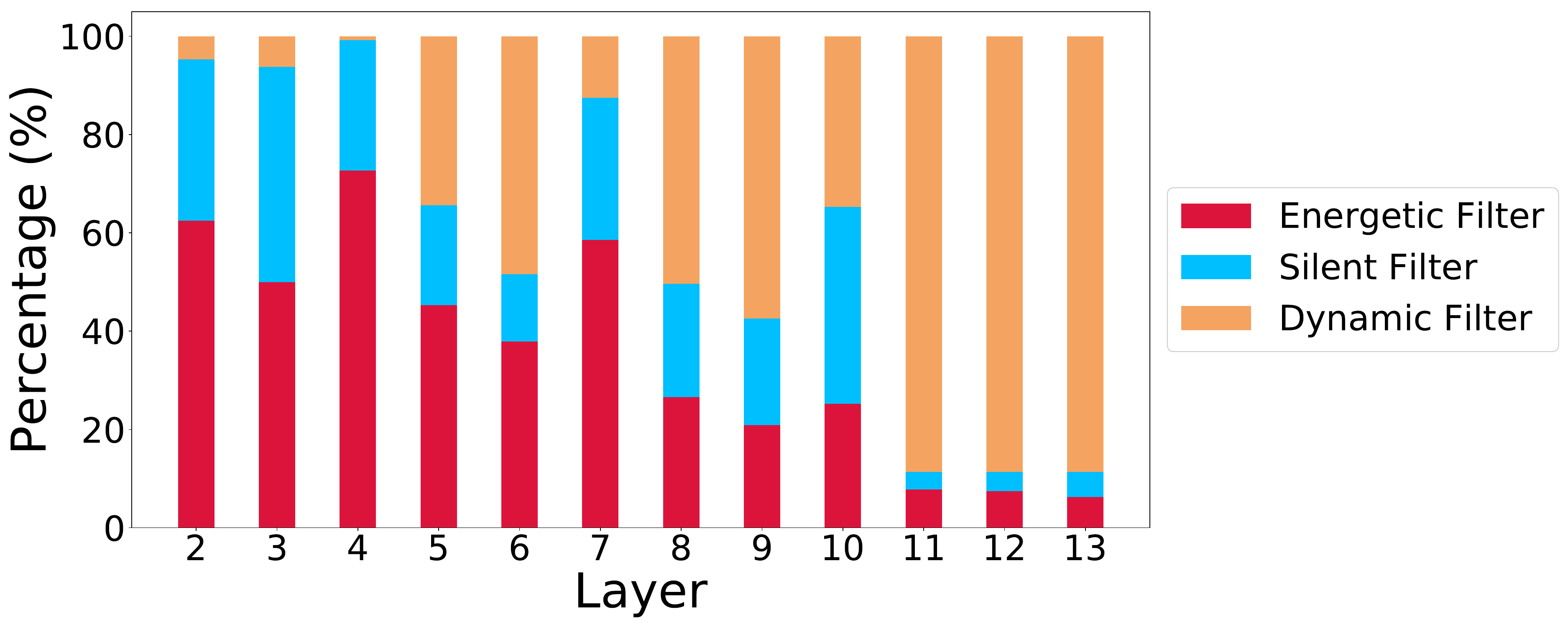}
\caption{The distribution of filters with different states in each layer of VGG-16 on ImageNet2012.}
\label{fig:layer_sum}
\vspace{-1.5em}
\end{wrapfigure}
\textbf{Filter State.}
After decoupling the network architecture, the state of a filter in the network has three possibilities: it responds to all the input samples, it does not respond to any input samples, or it responds to the specific inputs.
These three possibilities are termed as \textit{energetic filter}, \textit{silent filter}, and \textit{dynamic filter}, respectively.
As shown in Fig.~\ref{fig:layer_sum}, we collect different states of filters in different layers.
We can see that the proportion of dynamic filters increases with network depth increasing.
This phenomenon demonstrates that filters in the top layer tend to detect high-level semantic features, which are highly related to the input images.
In contrast, filters in the bottom layer tend to detect low-level features, which are always shared across images.

\begin{figure}
\begin{center}
\includegraphics[width=0.5\columnwidth]{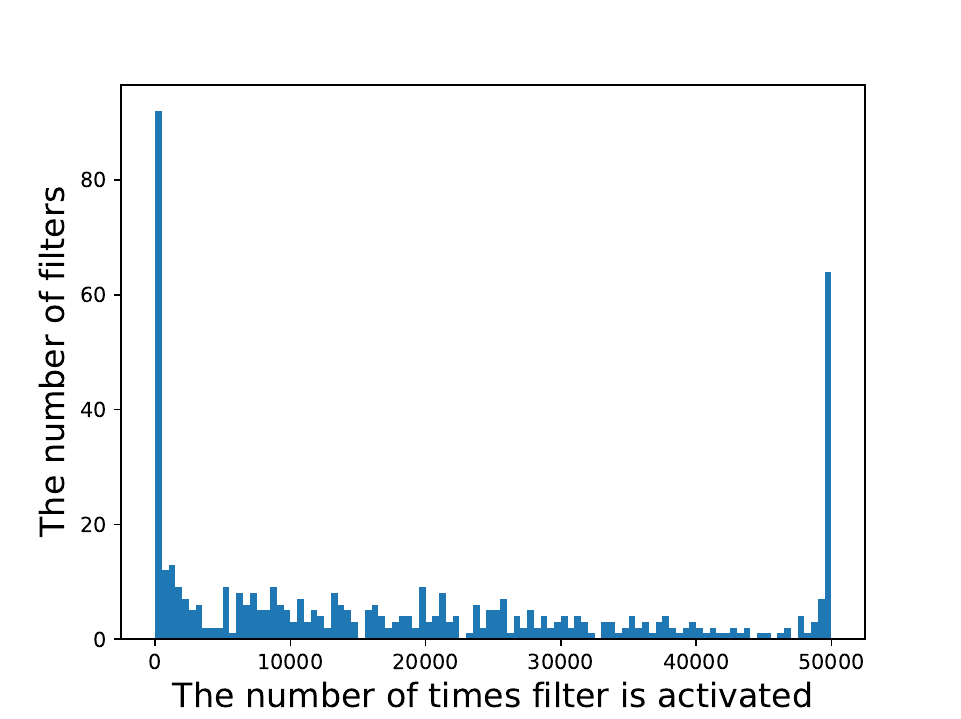}
\end{center}
  \vspace{-1em}
    \caption{The distribution of the number of times filters activated on ImageNet 2012 validation set. These filters are collected from the last convolutional layer of VGG-16.}
    \vspace{-1em}
    \label{fig:filter_num}
\end{figure}

As shown in Fig~\ref{fig:filter_num}, we collect filters in the last convolutional layer of VGG16 on ImageNet 2012 after network decoupling and present the results of the number of times they are activated.
ImageNet 2012 contains 50,000 validation images, so the number of times each filter is activated falls in the interval $[0, 50000]$.
The leftmost bar represents the number of filters that never been activated (\emph{i.e.,} silent filters), and the rightmost bar represents the number of filters that are activated every time (\emph{i.e.,} energetic filters).
The middle bars represent the number of filters, which respond to specific inputs (\emph{i.e.,} dynamic filters).
For instance, the rightmost bar represents that there are 60 filters, which are activated 50,000 times during evaluating on ImageNet 2012, in the last convolutional layer of VGG-16 after network decoupling.
They represent three different roles played by filters in the network:
Silent filters represent the redundant information, dynamic filters are responsible for specific semantic concepts.
A special case is the energetic filters, the existence of which attributes to the fact that most networks are limited in width (\emph{i.e.,} the number of filters).
the networks need some energetic filters which encode the more semantic concepts rather than a specific one.
After that, energetic filters are participated in the calculation path of all input images to improve the network performance.

\begin{table}[t]
 	\begin{tabular}{p{2.7cm}<{\centering}|p{1.5cm}<{\centering}|p{1.5cm}<{\centering}|p{1.4cm}<{\centering}|p{1.4cm}<{\centering}|p{1.4cm}<{\centering}|p{1.4cm}<{\centering}}
  \hline
  Model & Top1-Acc & Top5-Acc & Conv2\_2 & Conv3\_3 & Conv4\_3 & Conv5\_3 \\
   \hline
   VGG-16 & 71.59 & 90.38 & 0.0637 & 0.0446 & 0.0627 & 0.0787 \\
   VGG-16$_{decoupled}$ & 71.51 & 90.32 & \textbf{0.0750} & \textbf{0.0669} & \textbf{0.0643} & \textbf{0.0879} \\
   \hline
   Model & Top1-Acc & Top5-Acc & Block1 & Block2 & Block3 & Block4 \\
   \hline
   ResNet-18 & 69.76 & 89.08 &0.0527 & 0.0212 & 0.0477 & 0.0521 \\
   ResNet-18$_{decoupled}$ & 67.62 & 87.78 & \textbf{0.1062} & \textbf{0.0268} & \textbf{0.0580} & \textbf{0.0618} \\
   \hline
\end{tabular}
\vspace{0.3em}
\makeatletter\def\@captype{table}\makeatother\caption{The average interpretability score of filters in the different layers of original networks and decoupled networks on BRODEN. The higher score is better.}
\vspace{-2em}
\label{tab:filter_inter}
\end{table}
\begin{figure}[t]
  \centering
    \subfigure[color] {
    \label{fig:resnet18_color}
    \includegraphics[width=0.3\columnwidth]{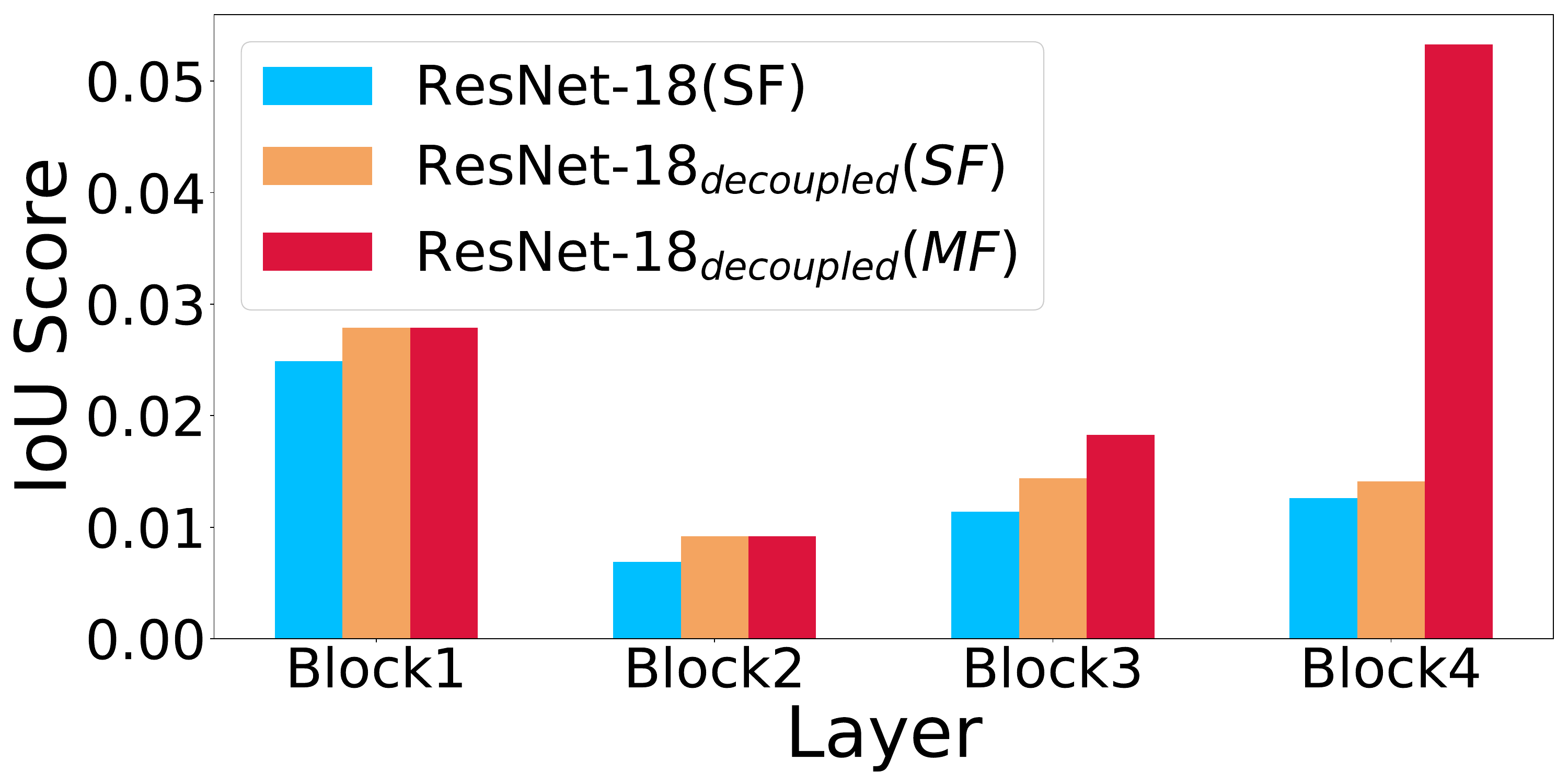}
  }
  \subfigure[objects] {
  \label{fig:resnet18_object}
  \includegraphics[width=0.3\columnwidth]{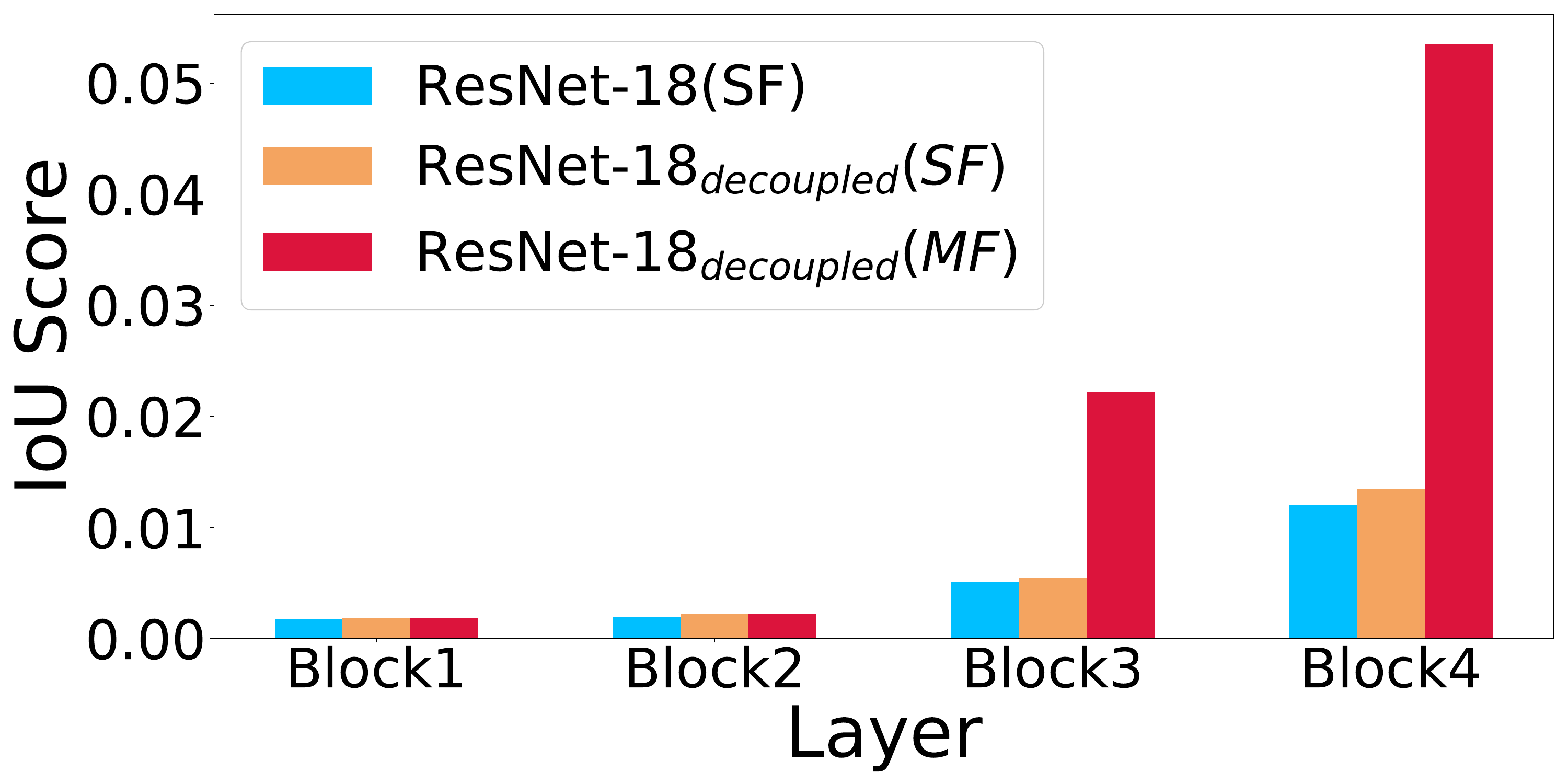}
  }
  \subfigure[parts] {
  \label{fig:resnet18_part}
  \includegraphics[width=0.3\columnwidth]{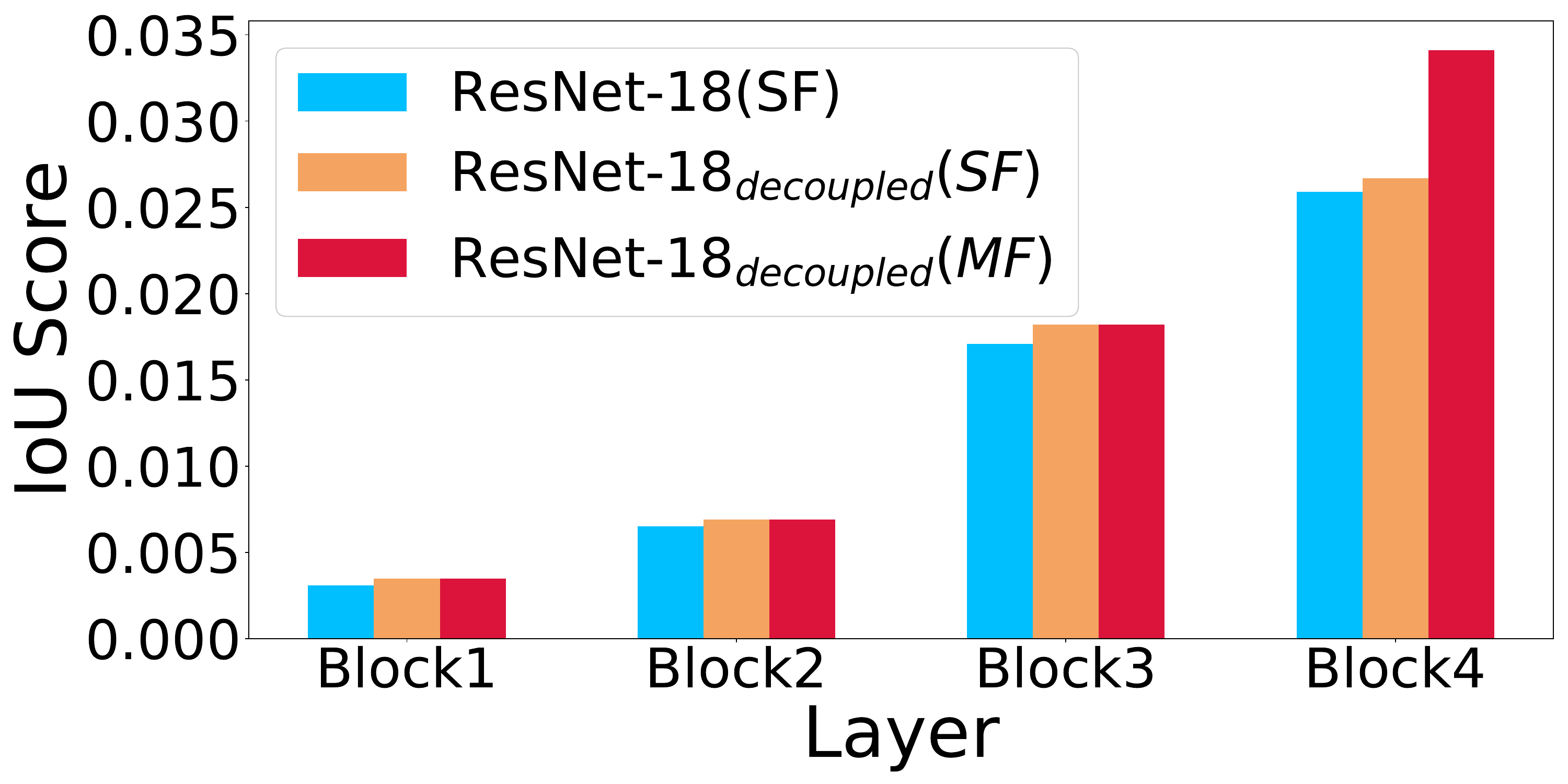}
  }
  \subfigure[materials] {
  \label{fig:resnet18_material}
  \includegraphics[width=0.3\columnwidth]{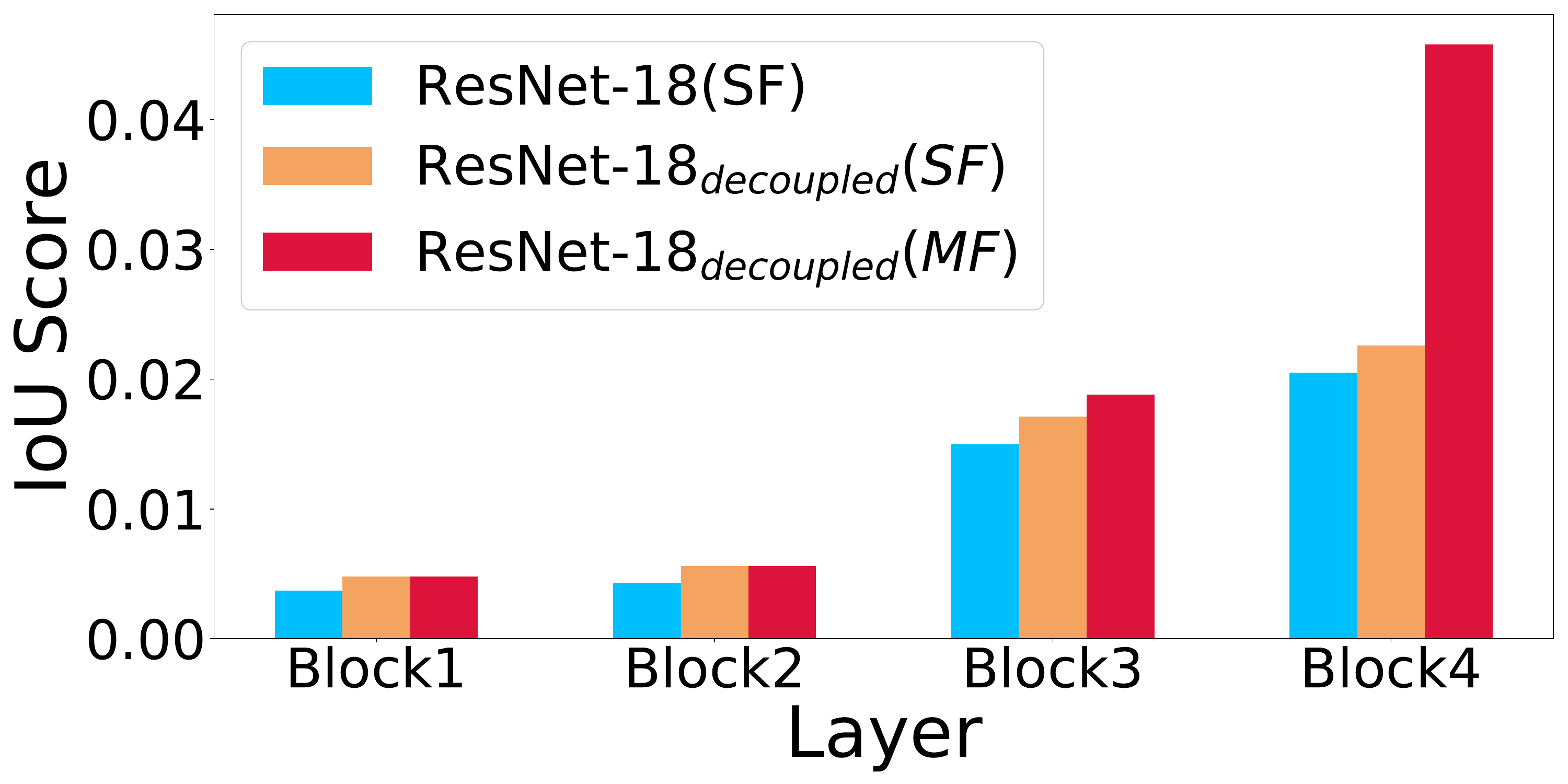}
  }
  \subfigure[scenes] {
  \label{fig:resnet18_scene}
  \includegraphics[width=0.3\columnwidth]{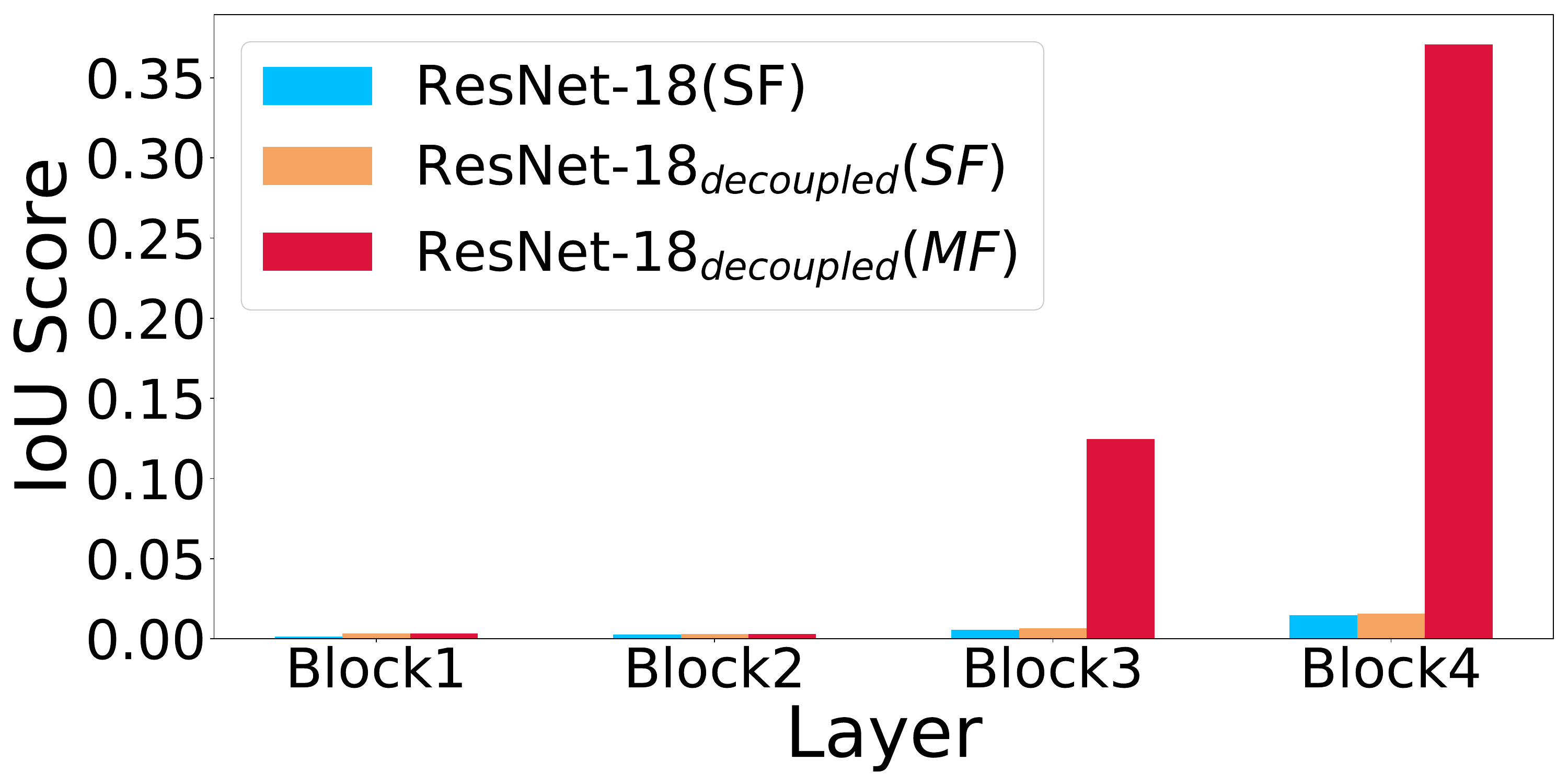}
  }
  \subfigure[textures] {
  \label{fig:resnet18_texture}
  \includegraphics[width=0.3\columnwidth]{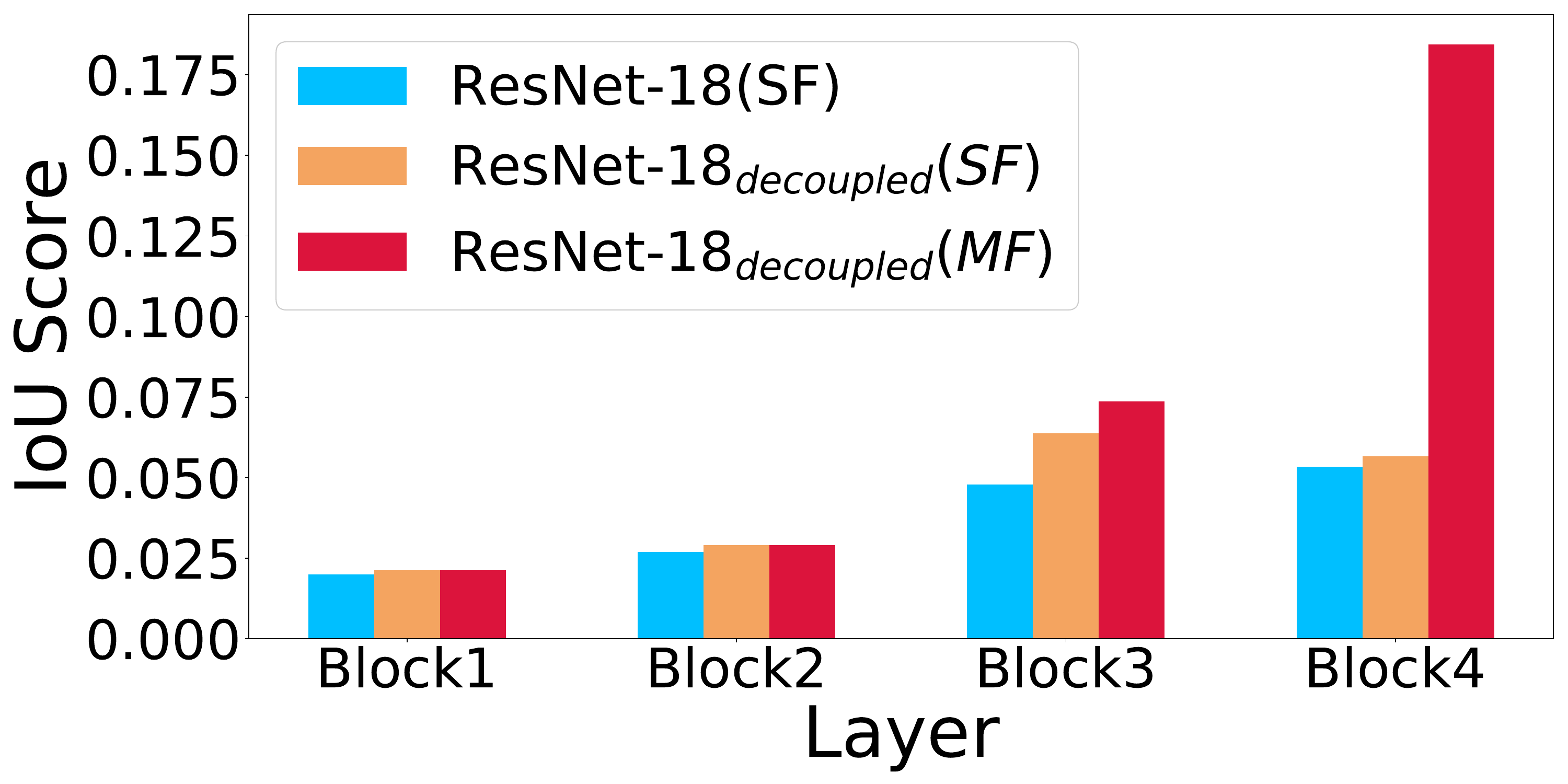}
  }
  \vspace{-1.5em}
     \caption{Average representation ability of different concepts in ResNet-18 on BRODEN. SF/MF represents use single/multiple filters characterizing the semantic features, respectively.}
  \label{fig:interpretability_metric}
       \vspace{-2em}
  \end{figure}

\textbf{Interpretable Quantitative Analysis.}
Following the works \cite{bau2017network,zhang2018interpretable,fong2018net2vec}, we select the interpretability of filters and the representation ability of semantic features to measure the network interpretability.
Specifically, we first select the original and our decoupled models which trained on ImageNet2012, and compute the activation map of each filter/unit on BRODEN dataset.
Then, the top quantile level threshold is determined over all spatial locations of feature maps.
After that, low-resolution activation maps of all filters are scaled up to input-image resolution using bilinear interpolation and thresholded into a binary segmentation, so as to obtain the receptive fields of filters.
The score of each filter $f$ as segmentation for the semantic concept $t$ in the input image $I$ is reported as an intersection-over-union score $IoU^I_{f,t} = \frac{|S_f^I \cap S_t^I|}{|S_f^I \cup S_t^I|}$, where $S_f^I$ and $S_t^I$ denote the receptive field of filter $f$ and the ground-truth mask of the semantic concept $t$ in the input image, respectively.
Given an image $I$, we associated filter $f$ with the $t$-th part if $IoU^I_{f,t} > 0.01$.
Finally, we measure the relationship between the filter $f$ and concept $t$ by $P_{f, t} = mean_{I}\textbf{1}(IoU^I_{f,t} > 0.01)$ across all the input images. 
Based on \cite{zhang2018interpretable}, we can report the highest association between the filter and concept as the final interpretability score of filter $f$ by $max_{t} P_{f, t}$.
As shown in Table. \ref{tab:filter_inter}, the value in each layer is obtained by averaging the final interpretability score across all the corresponding filters.
For ResNet-18, we collect the filters from the first convolutional layers in the last unit of each block.
Compared to the original networks, our decoupled networks have the better interpretability under the similar classification accuracy.
For instance, we achieve $1.2\times \sim 2\times$ score improvement of the filter interpretability than the original ResNet-18.

 \begin{figure*}[t]
  \centering
    \includegraphics[width=1\columnwidth]{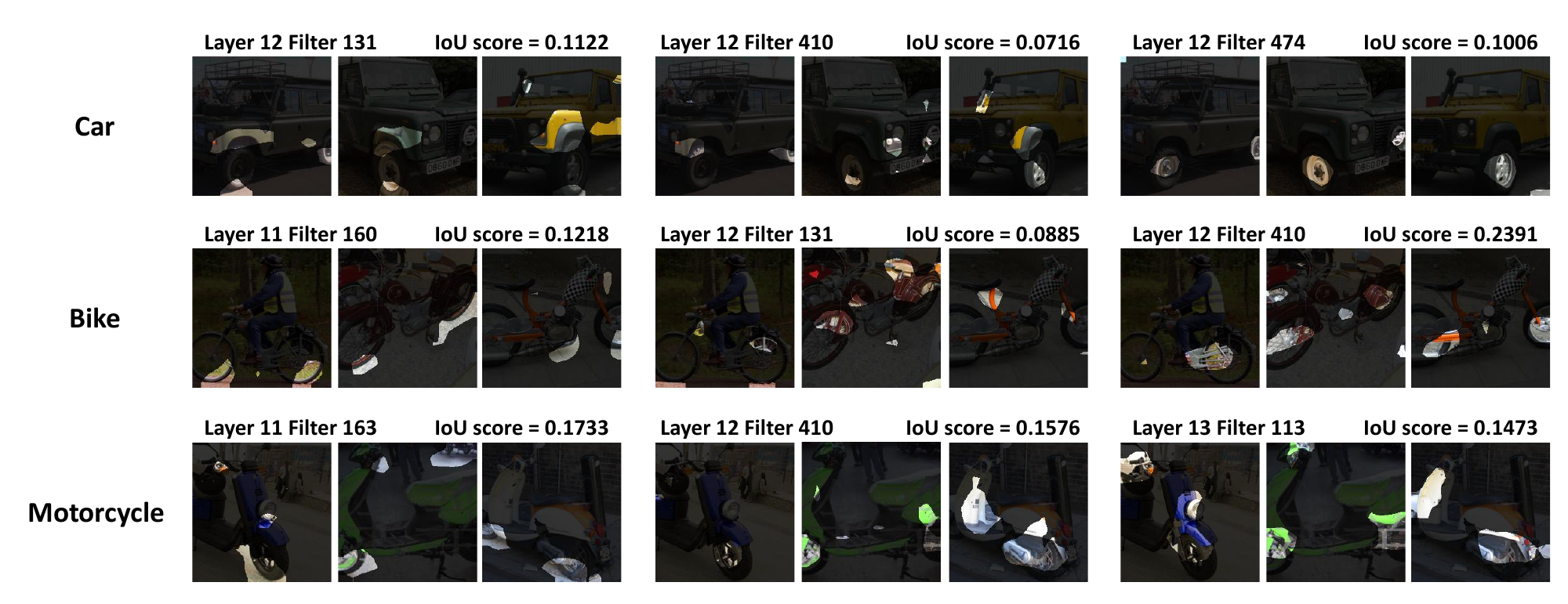}
  \vspace{-2.5em}
      \caption{Visualization of the receptive fields of filters which are inactivated because of the lack of semantic feature in images. We occlude the specific semantic feature (\emph{i.e.,} wheel) in different images (\emph{i.e.,} car, bike and motorcycle) on ImageNet and then collect the filters become inactivated due to the lack of the semantic feature.}
      \label{fig:path_1}
      \vspace{-1.8em} 
  \end{figure*}
  
  \begin{figure*}
  \begin{center}
    \includegraphics[width=1\columnwidth]{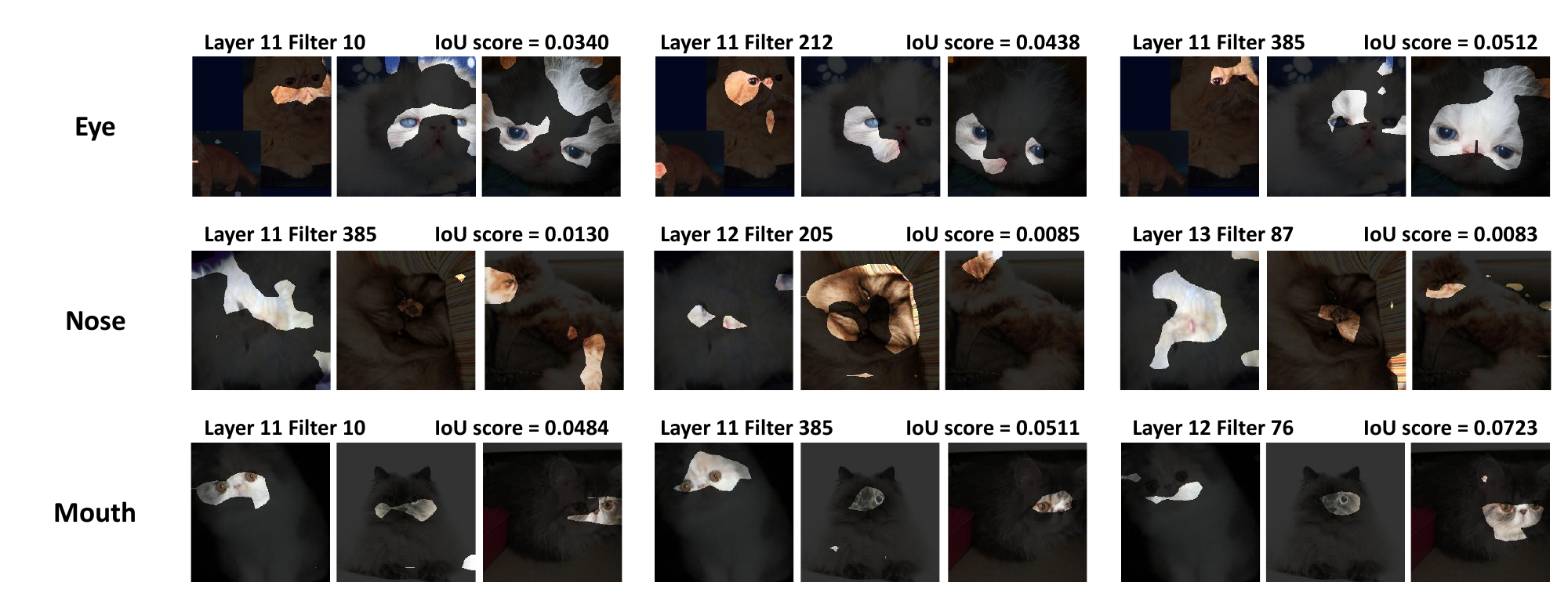}
  \end{center}
  \vspace{-.5em}
      \caption{Visualization of the receptive fields of filters which are inactivated because of the lack of semantic feature (\emph{i.e.,} eye, nose and mouth) in images (\emph{i.e.,} cat).}
      \label{fig:path_2}
      \vspace{-1em} 
  \end{figure*}

We further investigate the representation ability of network for specific semantic features before and after network decoupling.
For the representation of semantic features from a single filter, we evaluate the highest association between each semantic feature in BRODEN (which has $1,197$ semantic features) and the filters using $max_{I, f} IoU^I_{f,t}$ as the representation ability of specific semantic features, based on \cite{fong2018net2vec}.
For the representation of semantic features from multiple filters, we first occlude the semantic features in the original image and then collect the number of $M$ filters by comparing the difference between the calculation path of the original image and the occluded image, where these filters are activated on the original image but inactivated due to the lack of specific semantic features.
%
%
After that, we merge their receptive field and calculate the value of IoU $IoU^I_{f \in M,t} = \frac{|S_{f \in M}^I \cap S_t^I|}{|S_{f \in M}^I \cup S_t^I|}$ as the representation ability of semantic feature $t$.
As shown in Fig. \ref{fig:interpretability_metric}, we average the representation ability of semantic features belonging to the same concepts in the different layers.
The results demonstrates that our decoupled network has the better representation ability of semantic feature than the original ResNet-18.
The combination of multiple filters, which collected by our path-level disentangling, achieves about $3 \times$ improvement in the representation ability than the single ones.
Moreover, we find that the bottom layers in the decoupled network always use the single filters to characterize the semantic features based on our path-level analysis, so the representation ability of semantic features in the bottom layers is similar in the single filter and multiple filters.

\begin{figure*}[t]
\centering
  \hspace{-1em}
    \subfigure[Different kinds of objects.]{
      \label{fig:all_path}
      \includegraphics[width=0.49\columnwidth]{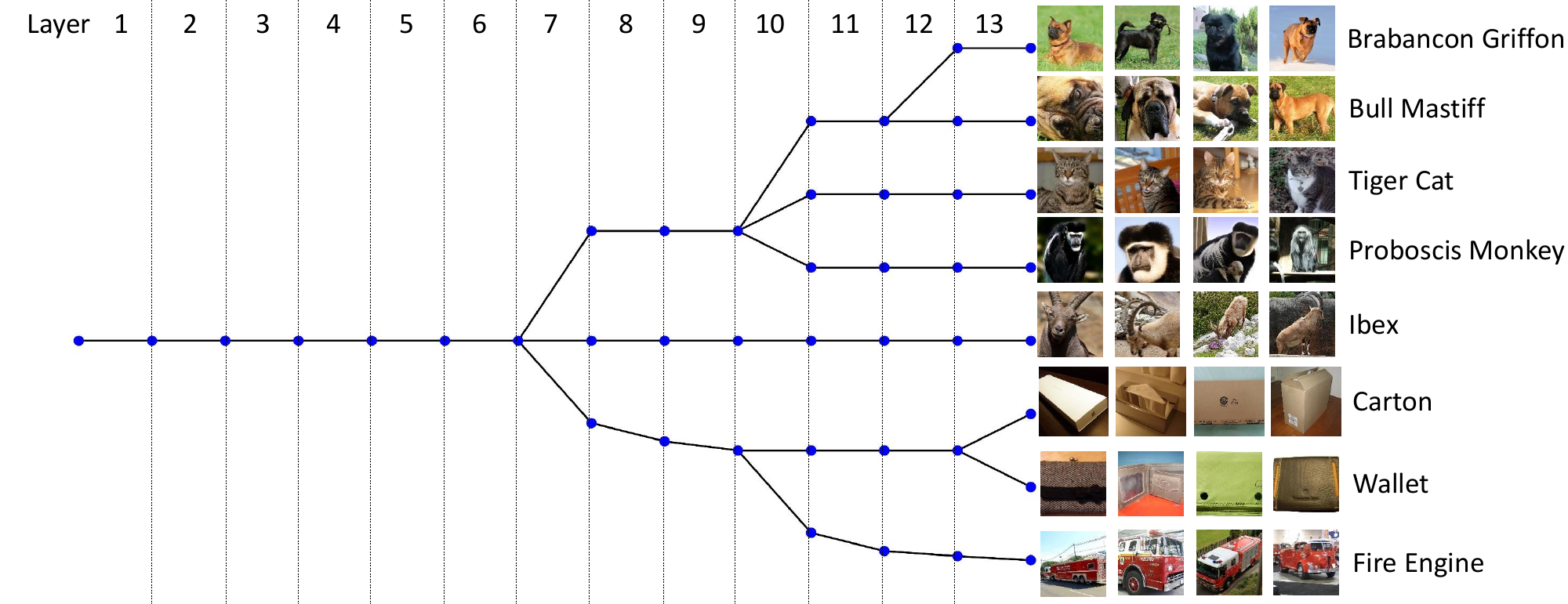}
    }
    \hspace{-1em}
    \subfigure[Fine-grained dogs.]{
      \label{fig:dog_path}
      \includegraphics[width=0.49\columnwidth]{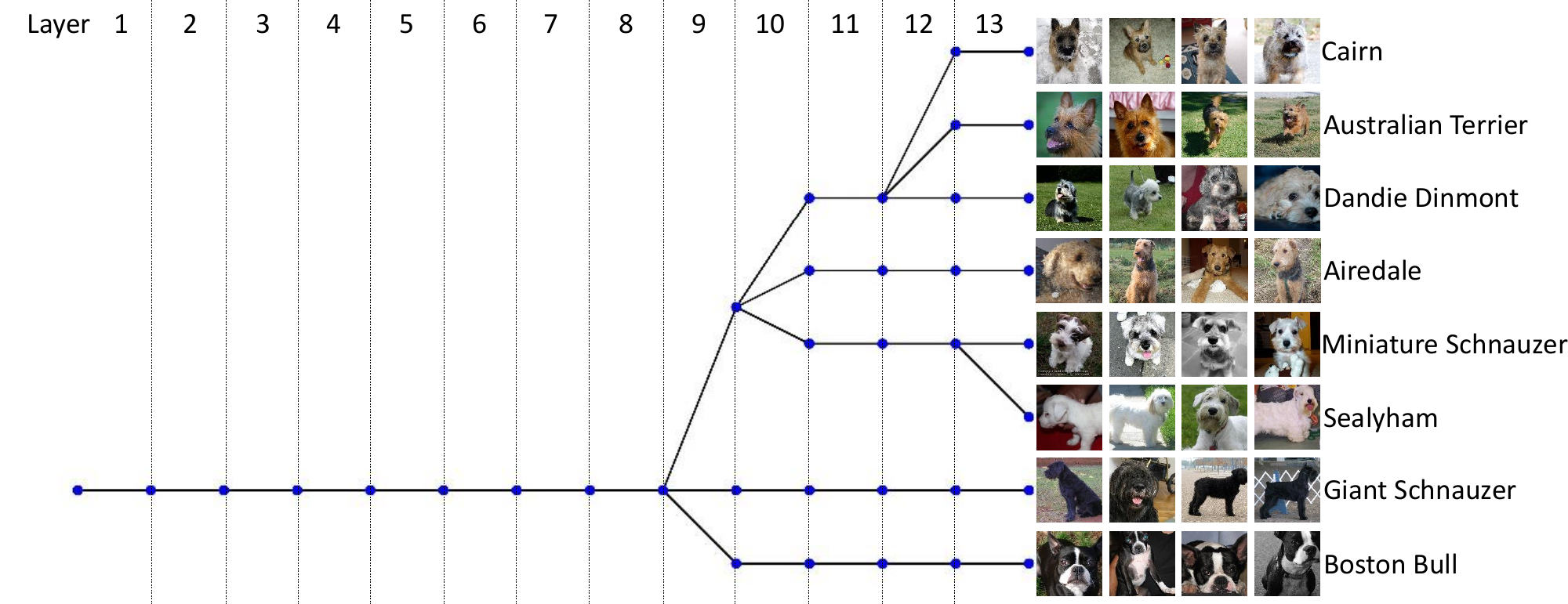}
    }
  \vspace{-1.2em}
      \caption{Visualization of the decision-making process of VGG-16 based on eight different categories of images.}
      \label{fig:class_path}
      \vspace{-1.8em} 
  \end{figure*}

\textbf{Semantic Concept Analysis.}
We further investigate the relationship between semantic concepts and calculation paths.
To this end, we occlude the areas that contain similar semantic features (\emph{i.e.,} wheels) in the images from different categories (\emph{i.e,.} car, bike and motorcycle) to analyze the characterization of the same semantic concept in different categories.
After that, we collect the filters which in the different parts of calculation path between the original images and the semantic lacked images.
Our experiments only collect the three filters with highest IoU score in the last three convolutional layers of VGG-16.
We find that the existence of a single semantic concept affects the state of multiple filters.
For example, as shown in the first row of Fig.~\ref{fig:path_1}, when we only occlude the wheels of the car with black blocks, the $131$-th, $410$-th and $474$-th filters in the $12$-th convolutional layer become inactived, which makes the calculation path change.
To further analyze the relationship between each filter and semantic concept, we visualize the receptive fields of filters on the input image to obtain the specific detection location of each one, and calculate the IoU score between the receptive fields of filters and the location area of the semantic concept.
We find that different filters are responsible for different parts of the same semantic concept.
For instance, the $131$-th, $410$-th and $474$-th filters in the $12$-th convolutional layer of VGG-16 are responsible for the features in the different parts of the wheel in ``car'' images, respectively.
Therefore, the combination of these filters has the better representation ability of the wheel than the single ones. 
%
Moreover, when the wheel is occluded in different images, the $410$-th filter in the $12$-th convolutional layer is always inactived because this filter is responsible for the texture and shape of the wheel, which is similar in different types of vehicles.

To explore the relationship between the characterization of different semantic concepts in the network, we also visualize the different semantic concepts for the same category of images, as shown in Fig.~\ref{fig:path_2}.
The different parts of the cats (\emph{i.e.,} eye, nose and mouth) are occluded by black blocks, and then the filters becomes inactivated due to the lack of semantic features have been collected.
As shown in the Fig.~\ref{fig:path_2}, the $385$-th filter in the $11$-th convolutional layer of VGG16 is always inactivated due to the lack of some features of cat.
This demonstrates that the semantic concepts detected by this filter are covering the entire cat face, including eye, nose and mouth.
Other filters are only responsible for a single semantic concept.
For example, the $76$-th and $205$-th filters in the $12$-th convolutional layer only detects the mouth and nose of the cats, respectively.

\textbf{Decision-Making Process of a Network.}
To investigate the decision-making process of a network and the functional process of its intermediate layers, we collect the calculation paths of a decoupled VGG-16 from eight different categories of images, which contain different kinds of artifacts and animals in Fig.~\ref{fig:all_path}, and fine-grained dogs in Fig.~\ref{fig:dog_path}.
We first collect the architecture encoding vectors layer-by-layer and then compute their Hopkins Statistic \cite{hopkins1954new} to analyze whether the inputs have different calculation paths in this layer.
If yes, we divide inputs into two subclasses by k-means.
In contrast, we keep the inputs in the same class and turn to the next layer.
The results show that the bottom layers in the network are responsible for general features, thus all inputs share the same calculation path.
As shown in Fig.~\ref{fig:all_path}, the decoupled VGG-16 cannot distinguish the difference between artifacts and animals until reaching the $7$-th convolutional layer.
Moreover, the network distinguishes the difference in the fine-grained dogs after reaching the $9$-th convolutional layer, as shown in Fig.~\ref{fig:dog_path}.
As the layers become deeper, the network gradually distinguishes the different objects, and similar objects are distinguished in the last layers.

\begin{table*}[t]
\begin{minipage}[b]{0.51\textwidth} 
\begin{tabular}{p{1.7cm}<{\centering}|p{1cm}<{\centering}|p{1cm}<{\centering}|p{1cm}<{\centering}|p{1cm}<{\centering}}
  \hline
  \multirow{2}*{Model} & \multicolumn{2}{c|}{CIFAR-10} & \multicolumn{2}{c}{CIFAR-100} \\

  \cline{2-5}

  ~ & \multirow{2}*{FLOPs}  &Top-1 Acc(\%) & \multirow{2}*{FLOPs} & Top-1 Acc(\%) \\
  \hline

  ResNet-56 & 125M & 93.17 & 125M & 70.43 \\
  \hline

  CP \cite{he2017channel} & 63M & 91.80 & - & - \\
  \hline

  L1 \cite{li2016pruning}$^*$ & 90M & 93.06 & 86M & 69.38 \\
  \hline

  Skip \cite{wang2018skipnet}$^*$ & 103M & 92.50 & - & - \\
  \hline

  \textbf{Ours} & \textbf{63M} & \textbf{93.08} & \textbf{41M} & \textbf{69.72} \\
  
  \hline\hline

  VGGNet & 398M & 93.75 & 398M & 72.98 \\
  \hline

  L1 \cite{li2016pruning}$^*$ & 199M & 93.69 & 194M & 72.14 \\
  \hline

  Slim \cite{liu2017learning} & 196M & 93.80 & 250M & 73.48 \\
  \hline

  \textbf{Ours} & \textbf{141M} & \textbf{93.82} & \textbf{191M} & \textbf{73.84} \\
  \hline\hline

  GoogleNet & 1.52B & 95.11 & 1.52B & 77.99 \\
  \hline

  L1 \cite{li2016pruning}$^*$ & 1.02B & 94.54 & 0.87B & 77.09 \\
  \hline
  
  \textbf{Ours} & \textbf{0.39B} & \textbf{94.65} & \textbf{0.75B} & \textbf{77.28} \\
  \hline

\end{tabular}
 \vspace{0.5em}
\caption{Results of the different networks on CIFAR-10 and CIFAR-100. $*$ represents the result based on our implementation.}
\vspace{-1em}
\label{tab:cifar10}

  \end{minipage}
\hspace{0.2em}
\vspace{-1.4em}
\begin{minipage}[b]{0.45\textwidth} 
\begin{tabular}{p{1.33cm}<{\centering}|p{1cm}<{\centering}|p{1cm}<{\centering}|p{1cm}<{\centering}|p{1cm}<{\centering}}
  \hline
  \multirow{2}*{Model} &  Top-1 Acc$\downarrow$ (\%) & Top-5  Acc$\downarrow$ (\%) & FLOPs Reduction & CPU Time Reduction\\
  \hline

  SFP \cite{he2018soft} & 3.18 & 1.85 & 1.72$\times$ & 1.38$\times$ \\
  \hline

  DCP \cite{zhuang2018discrimination} & 2.29 & 1.38 & 1.89$\times$ & 1.60$\times$ \\
  \hline
  
  LCL \cite{dong2017more} & 3.65 & 2.30 & 1.53$\times$ & 1.25$\times$\\
  \hline

  FBS \cite{gao2018dynamic} & 2.54 & 1.46 & 1.98$\times$ & 1.60$\times$ \\
  \hline

  \textbf{Ours} & \textbf{2.14} & \textbf{1.30} & \textbf{2.03$\times$} & \textbf{1.64$\times$}\\
  \hline

\end{tabular}
 \vspace{0.3em}
\caption{Results of ResNet-18 on ImageNet2012. The baseline in our method has an 69.76\% top-1 accuracy and 89.08\% top-5 accuracy with $1.81$B FLOPs and an average $180$ ms testing on CPU based an image by running the whole of the validation dataset.}
\label{tab:imagenet}
  \end{minipage}
\end{table*}

\begin{table*}
    \centering
  \begin{tabular}{p{5cm}<{\centering}|p{1.5cm}<{\centering}|p{1.5cm}<{\centering}|p{2cm}<{\centering}|p{2cm}<{\centering}}
    \hline
    \multirow{2}*{Model} &  Top-1 Acc(\%) & Top-5  Acc(\%) & FLOPs Reduction & CPU Time Reduction\\
    \hline
    
    Perforated CNNs \cite{figurnov2016perforatedcnns} & - & 88.8 & - & 2.00$\times$ \\
    \hline

    RunTime Neural Pruning \cite{lin2017runtime} & - & 87.58 & 3.00$\times$ & - \\ 
    \hline
  
    ThiNet \cite{luo2017thinet} & 69.80 & 89.53 & 3.23$\times$ & - \\
    \hline
    
    Global and Dynamic Filter Pruning \cite{lin2018accelerating} & 68.80 & 88.77 & 2.42$\times$ & 1.62$\times$ \\
    \hline
  
    Feature Boosting and Suppression \cite{gao2018dynamic} & - & 89.86 & 3.00$\times$ & \textbf{2.97$\times$} \\
    \hline
  
    \textbf{Decoupling} & \textbf{71.51} & \textbf{90.32} & \textbf{3.23$\times$} & 2.44$\times$\\
    \hline
  
  \end{tabular}
   \vspace{0.3em}
  \caption{Results of VGG-16 on ImageNet2012. The baseline in our method has $15.48$B FLOPs and an average $1220$ ms testing on CPU based an image by running the whole of the validation dataset.}
  \label{tab:imagenet_vgg}
  \vspace{-1em}
  \end{table*}

\vspace{-1em}
\subsection{Network Acceleration}

In this subsection, we evaluate how our method can facilitate network acceleration. 
We decouple three different network architectures (\emph{i.e.,} ResNet-56, VGGNet and GoogleNet) on CIFAR-10 and CIFAR-100, and set $R=0$ to allow the networks to be learned automatically.
The VGGNet in our experiments is the same as the network in \cite{liu2017learning}.
As shown in Table~\ref{tab:cifar10}, our method achieves the best trade-off between accuracy and speedup/compression rate, compared with static pruning \cite{he2017channel,li2016pruning,liu2017learning} and dynamic pruning \cite{wang2018skipnet}.
For instance, we achieve a $2\times$ FLOPs reduction with only a $0.09\%$ drop in top-1 accuracy for ResNet-56 on CIFAR-10.
For ImageNet 2012, the results of accelerating ResNet-18 are summarized in Table~\ref{tab:imagenet}.
When setting $R$ to $0.6$, our method also achieves the best performance with a $1.64\times$ real CPU running speedup and $2.03\times$ reduction in FLOPs compared with the static pruning \cite{he2018soft,zhuang2018discrimination} and dynamic pruning \cite{dong2017more,gao2018dynamic}, while only decreasing by $1.30$\% in top-5 accuracy.
For VGG-16 on ImageNet 2012, to achieve the best trade-off between accuracy and speed, we follow the \cite{luo2017thinet} to set $R=0.5$ in the first ten layers and $R=0.8$ in the last three layers.
As shown in Table~\ref{tab:imagenet_vgg}, we obtain 90.32\% Top-5 accuracy with a $2.44\times$ real CPU running speedup and $3.23\times$ reduction in FLOPs, which is better than static pruning \cite{figurnov2016perforatedcnns,luo2017thinet,lin2018accelerating} and dynamic pruning \cite{lin2017runtime,gao2018dynamic}.

The detailed of hyper-parameter settings in our experiments are shown in Table~\ref{tab:hyp}.
The $\lambda_m$, $\lambda_k$ and $\lambda_s$ control the influence of corresponding losses.
And the $R$ represents the target compression ratio.

\begin{table}
\centering
\begin{tabular}{p{3cm}<{\centering}|p{2cm}<{\centering}|p{2cm}<{\centering}|p{2cm}<{\centering}|p{2cm}<{\centering}}
  \hline
  Network &  $\lambda_m$ &  $\lambda_k$ & $\lambda_s$ & R\\
  \hline

  ResNet-56 & 0.01 & 1 & 0.00015 & 0 \\
  \hline

  VGGNet & 0.04 & 1 & 0.0002 & 0 \\
  \hline

  GoogleNet & 0.006 & 1 & 0.00005 & 0 \\
  \hline

  ResNet-18 & 0.005 & 1 & 0.01 & 0.6 \\
  \hline

  VGG-16 & 0.01 & 1 & 0.01 & 0.5/0.8 \\
  \hline

\end{tabular}
\vspace{0.5em}
\caption{Hyper-parameter settings on network acceleration.}
\label{tab:hyp}
\end{table}

    
\vspace{-1em}
\subsection{Adversarial Samples Detection}

\begin{figure}[htb]
\begin{minipage}[b]{0.5\textwidth} 
\centering
    \includegraphics[width=0.7\columnwidth]{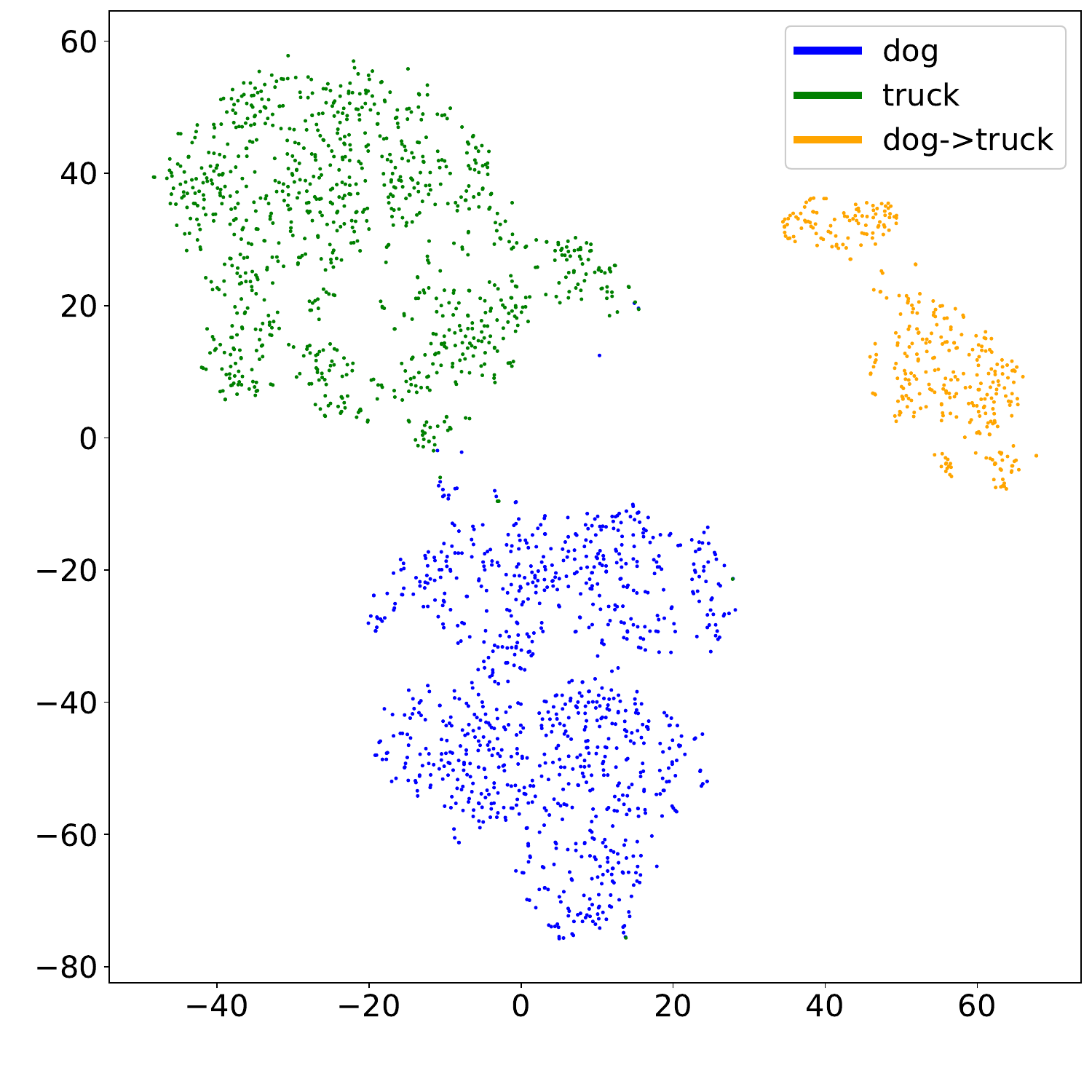}
\vspace{-2em} 
\caption{The distribution of the integral calculation path of original images and adversarial samples in ResNet-56 on CIFAR-10.}
      \label{fig:attack_change}
      \vspace{-1em} 
 \end{minipage}
 \hspace{0.5em} 
 \begin{minipage}[b]{0.45\textwidth}
 	\begin{tabular}{p{1.3cm}<{\centering}|p{1.1cm}<{\centering}|p{0.9cm}<{\centering}|p{0.9cm}<{\centering}|p{0.9cm}<{\centering}}
  \hline
  \multirow{2}*{Classifier} & \multirow{2}*{Method} & \multicolumn{3}{c}{Num. of samples} \\
  \cline{3-5}
  ~ & ~ & 1 & 5 & 10 \\
  \hline
 \multirow{2}*{\tabincell{c}{random \\ forest}} &\cite{wang2018interpret} & 0.879 & 0.894 & 0.904 \\ 
   \cline{2-5}
 ~ & Ours & \textbf{0.903} & \textbf{0.941} & \textbf{0.953} \\
  \hline
  \multirow{2}*{adaboost} &\cite{wang2018interpret} & 0.887 & 0.905 & 0.910 \\ 
   \cline{2-5}
 ~ & Ours & \textbf{0.909} & \textbf{0.931} & \textbf{0.940} \\
 \hline
 \multirow{2}*{\tabincell{c}{gradient \\ boosting}} &\cite{wang2018interpret} & 0.905 & 0.919 & 0.915 \\ 
   \cline{2-5}
 ~ & Ours & \textbf{0.927} & \textbf{0.921} & \textbf{0.928} \\
 \hline
  
\end{tabular}
\makeatletter\def\@captype{table}\makeatother\caption{The Area-Under-Curve (AUC) score on adversarial samples detection. Higher is better.}
\label{tab:attack_auc}

 \end{minipage}

  \end{figure}

We further demonstrate that the proposed architecture decoupling can help to detect the adversarial samples.
Recently, several works \cite{goodfellow2014explaining} have concluded that neural networks are vulnerable to adversarial examples, where adding a slight amount of noise to an input image can disturb their robustness.
We add noise to images belonging to the ``dog'' category to make the network predicts as ``truck'' and visualize the distribution of the calculation path between the original images and adversarial samples in ResNet-56 on CIFAR-10, as shown in Fig.~\ref{fig:attack_change}.
The result demonstrates that the calculation path of the adversarial samples ``dog$\to$truck'' is different from that of the original ``dog'' and ``truck'' images.
In other words, adversarial samples do not completely deceive our decoupled network, which can detect them by analyzing their calculation paths.
%

%
Based on the above observation, we use random forest, adaboost and gradient boosting as the binary classifier to determine whether the calculation paths are from real or adversarial samples.
As shown in Table~\ref{tab:attack_auc}, we randomly select 1, 5 and 10 images from each class in the ImageNet 2012 training set to organize three different scales training datasets.
The testing set is collected by selecting 1 image from each class in the ImageNet validation dataset.
Each experiment is run five times independently.
The results show that our method achieves an AUC score of 0.049 gain over Wang \emph{et al.} \cite{wang2018interpret} (\emph{i.e.,} 0.953 \emph{vs.} 0.904), when the number of training samples is 10 on random forest.
It also demonstrates that the calculation paths obtained by our method are better than Wang \emph{et al.} \cite{wang2018interpret}, with higher discriminability.

%

\begin{table*}
  \centering
\begin{tabular}{p{4.7cm}<{\centering}|p{1.5cm}<{\centering}|p{1cm}<{\centering}|p{1.5cm}<{\centering}|p{1.5cm}<{\centering}|p{1.5cm}<{\centering}}
  \hline
  \multirow{2}*{Method} & Top1-Acc(\%) & \multirow{2}*{FLOPs} & dynamic filters & slient filters & energetic filters \\
  \hline

  ResNet-56+ACM & 93.17 & 118M & 0.30\% & 13.25\% & 86.45\% \\
  \hline
  ResNet-56+ACM+$\mathcal{L}_{s}$ & 92.94 & 63M & 0.24\% & 59.60\% & 40.16\% \\
  \hline
  ResNet-56+ACM+$\mathcal{L}_{s}$+$\mathcal{L}_{kl}$ & 92.81 & 117M & 0.00\% & 15.22\% & 84.78\% \\
  \hline
  ResNet-56+ACM+$\mathcal{L}_{s}$+$\mathcal{L}_{mi}$ & 92.99 & 67M & 29.00\% & 32.55\% & 38.45\% \\
  \hline
  ResNet-56+ACM+$\mathcal{L}_{s}$+$\mathcal{L}_{mi}$+$\mathcal{L}_{kl}$ & 93.08 & 63M & 30.27\% & 35.24\% & 34.49\% \\
  \hline

\end{tabular}
 \vspace{0.3em}
\caption{Effect of the losses. ACM represents the architecture controlling module.}
\label{tab:loss}
\end{table*}

\begin{figure}
  \vspace{-1em}
  \begin{center}
    \subfigure[ResNet-56 without our losses.]{
      \label{fig:arch_org}
      \includegraphics[width=0.45\columnwidth]{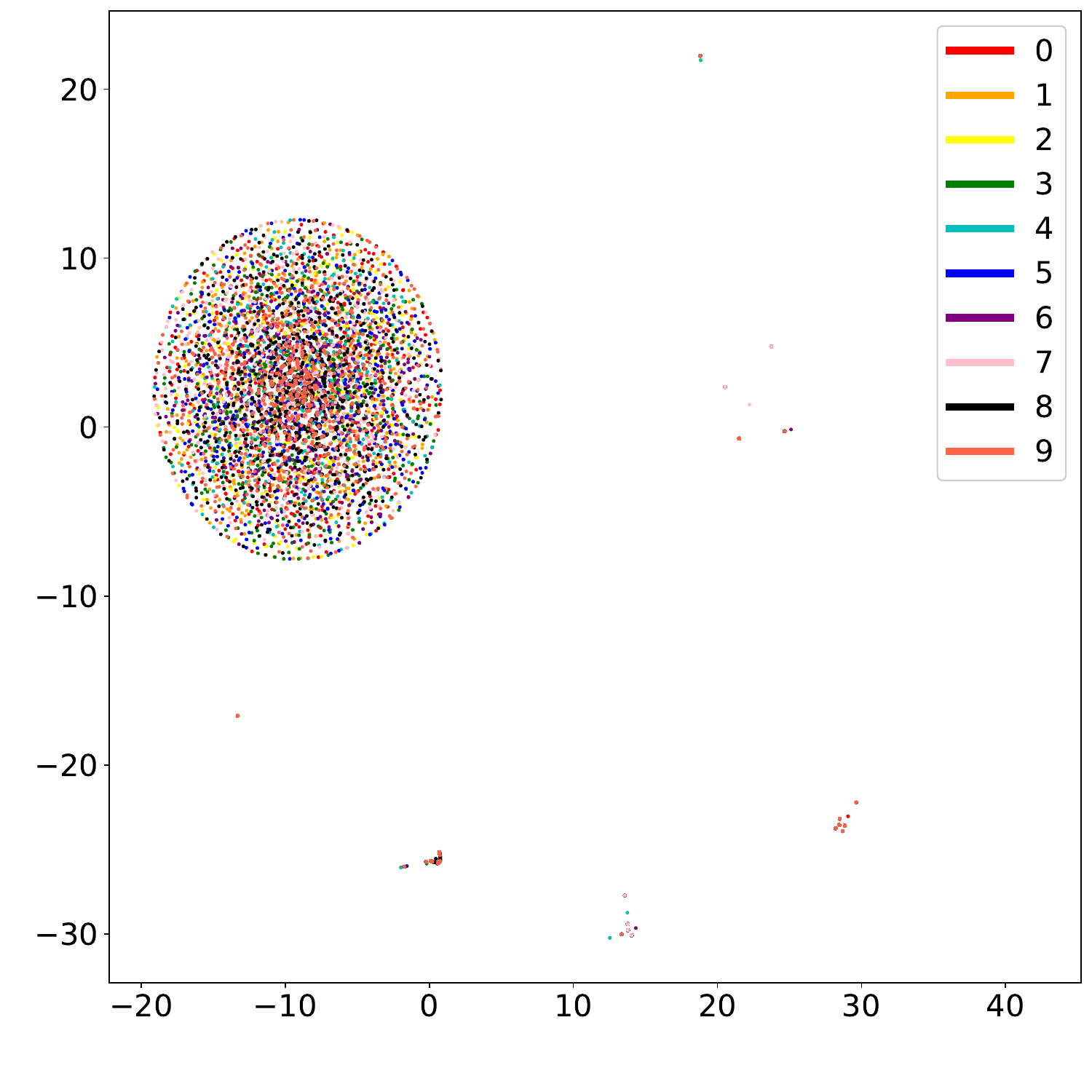}
    }
    \subfigure[ResNet-56 with our losses.]{
      \label{fig:arch_decoupled}
      \includegraphics[width=0.45\columnwidth]{arch_res56.pdf}
    }
  \end{center}
      \caption{Visualization of the distribution of the whole of calculation path in ResNet-56 on CIFAR-10 with or without losses proposed by our method.}
      \label{fig:decoupled}
  \end{figure}

\section{Ablation Study}

We train the ResNet-56 on CIFAR-10 with or without the losses proposed in our method to analyze the effect of the each loss.
As shown in Table~\ref{tab:loss}, the combination of three losses achieves the best trade-off between accuracy and FLOPs.
The lack of $\mathcal{L}_{mi}$ results in that the network tends to use static pruning to compress itself.
Meanwhile, compared with only using $\mathcal{L}_{s}$, using $\mathcal{L}_{kl}$ makes the filters respond to the all objects, which leads to the higher probability to generate the energetic filters.
Furthermore, as shown in Fig.~\ref{fig:decoupled}, after the training with the combination of these losses, we can decouple ResNet-56 successfully.

\begin{figure}
  \vspace{-1em}
  \begin{center}
    \subfigure {
    \label{fig:hyper_lambda_m}
    \includegraphics[width=0.3\columnwidth]{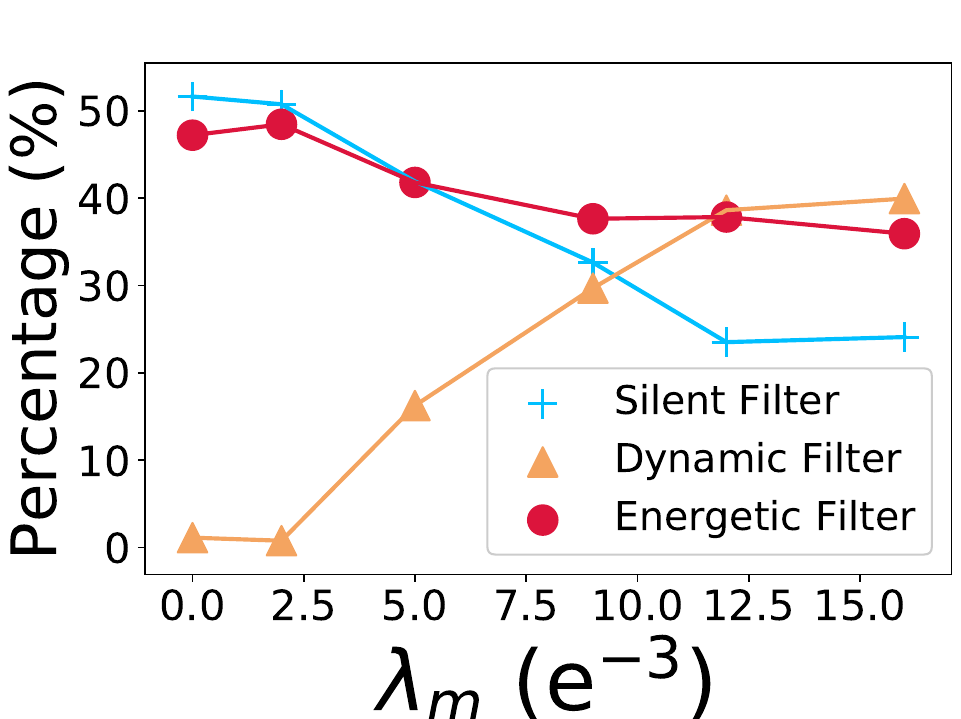}
    \vspace{-1em}
  }
  \subfigure {
  \label{fig:hyper_lambda_k}
  \includegraphics[width=0.3\columnwidth]{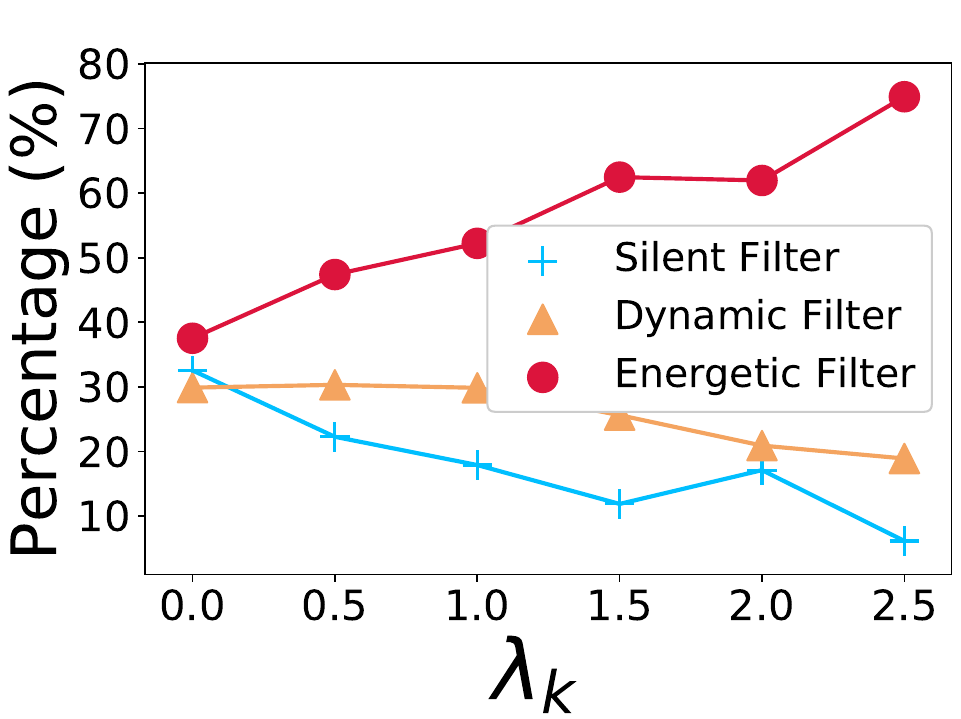}
  \vspace{-1em}
  }
  \subfigure {
  \label{fig:hyper_lambda_s}
  \includegraphics[width=0.3\columnwidth]{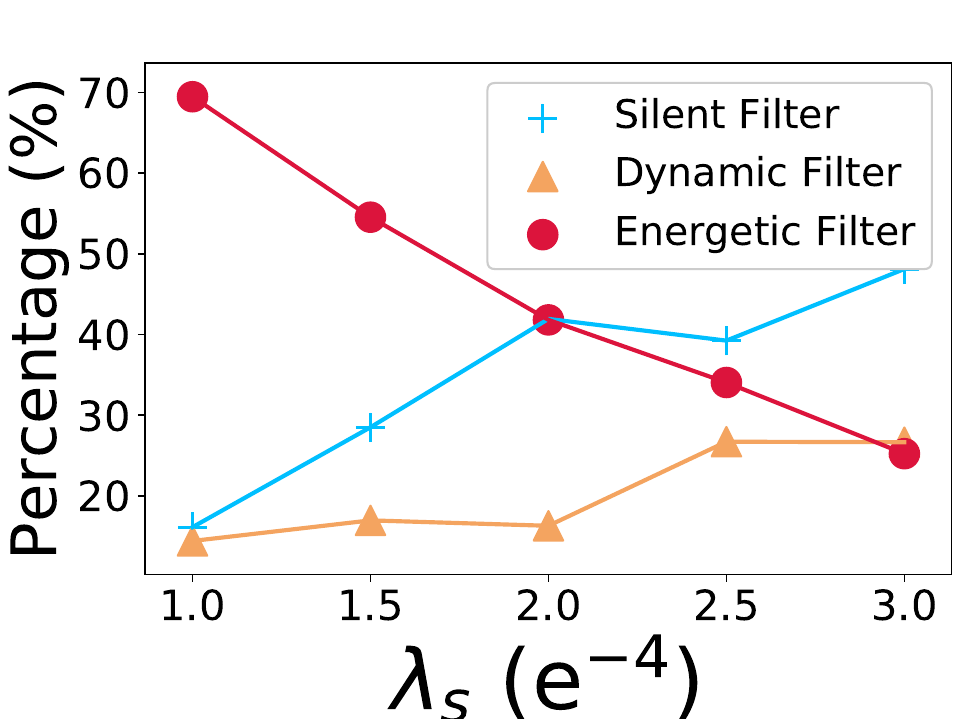}
  \vspace{-1em}
  }
  \end{center}
  \vspace{-1em}
     \caption{Percentages of different filter states for different $\lambda_m$, $\lambda_k$ and $\lambda_s$ on ResNet-56.}
  \label{fig:hyper_lambda}
  \vspace{-1em} 
  \end{figure}

Our decoupling method has three hyper-parameters (\emph{i.e.},  $\lambda_m$, $\lambda_k$ and $\lambda_s$) to control the network decoupling.
%
%
As shown in Fig.~\ref{fig:hyper_lambda}, we calculate the percentage of different filter states (\emph{i.e.,} energetic, silent and dynamic) in the network with different $\lambda_m$, $\lambda_k$ and $\lambda_s$ on ResNet-56.
We set $\lambda_m=0.01$, $\lambda_k=1$ and $\lambda_s=0.0001$ when they are not being measured.
We find that $\lambda_m$ controls the number of dynamic filters, which means the network architecture can be decoupled as $\lambda_m$ increases.
Meanwhile, $\lambda_s$ controls the number of filters that participate in the network inference, and $\lambda_k$ controls the number of energetic filters.

\vspace{-1.3em}
\section{Conclusion}
\vspace{-0.7em}
In this paper, we propose a novel architecture decoupling method to obtain an interpretable network and explore the rationale behind its overall working process based on a novel path-level analysis.
In particular, an architecture controlling module is introduced and embedded into each layer to dynamically identify the activated filters.
Then, by maximizing the mutual information between the architecture encoding vector and the input image, we decouple the network architecture to explore the functional processing behavior of each calculation path.
Meanwhile, to further improve the interpretability of the network and inference, we limit the output of the convolutional layers and sparsifying the calculation path.
Experiments show that our method can successfully decouple the network architecture with several merits, \emph{i.e.,} network interpretation, network acceleration and adversarial samples detection.

\textbf{Acknowledgements.} This work is supported by the Nature Science Foundation of China (No.U1705262, No.61772443, No.61572410, No.61802324 and No.61702136), National Key R\&D Program (No.2017YFC0113000, and No.2016Y
FB1001503), Key R\&D Program of Jiangxi Province (No. 20171ACH80022) and Natural Science Foundation of Guangdong Provice in China (No.2019B1515120049).

%
%
\bibliographystyle{splncs04}
\bibliography{egbib}
\end{document}